\pdfoutput=1
\documentclass[review]{nle}
\usepackage{url}
\usepackage[T1]{fontenc}
\usepackage{times,latexsym}
\usepackage{amsmath}
\usepackage{graphicx}
\usepackage{natbib}
\usepackage{multirow}
\usepackage{hyperref}
\usepackage{booktabs}
\usepackage{textcomp}
\usepackage{fixltx2e}
\usepackage[ruled,vlined]{algorithm2e}
\usepackage{algorithm2e}
\usepackage{flexisym}
\usepackage{blindtext}
\usepackage[none]{hyphenat}
\ifpdf%
\usepackage{epstopdf}%
\else%
\fi

\usepackage[ruled,vlined]{algorithm2e}
\usepackage{algorithm2e}

\begin{document}
\label{firstpage}

\lefttitle{PGST: a Polyglot Gender Style Transfer method}
\righttitle{Natural Language Engineering}
 
\papertitle{Article}

\jnlPage{1}{00}
\jnlDoiYr{2021}
\doival{10.1017/xxxxx}

\title{PGST: a Polyglot Gender Style Transfer method}

\begin{authgrp}
\author{Reza Khan Mohammadi}

\author{ Seyed Abolghasem Mirroshandel}
\affiliation{University of Guilan, Faculty of Engineering, \\
            Computer Engineering Department, Rasht, Iran
        \email{mirroshandel@guilan.ac.ir}}
\end{authgrp}
\begin{abstract}
Recent developments in Text Style Transfer have led this field to be more highlighted than ever. The task of transferring an input\textquotesingle{s} style to another is accompanied by plenty of challenges (e.g., fluency and content preservation) that need to be taken care of. In this research, we introduce PGST, a novel polyglot text style transfer approach in the gender domain, composed of different constitutive elements. In contrast to prior studies, it is feasible to apply a style transfer method in multiple languages by fulfilling our method\textquotesingle{s} predefined elements. We have proceeded with a pre-trained word embedding for token replacement purposes, a character-based token classifier for gender exchange purposes, and a beam search algorithm for extracting the most fluent combination. Since different approaches are introduced in our research, we determine a trade-off value for evaluating different models\textquotesingle{} success in faking our gender identification model with transferred text. To demonstrate our method\textquotesingle{s} multilingual applicability, we applied our method on both English and Persian corpora and ended up defeating our proposed gender identification model by 45.6\% and 39.2\%, respectively. While this research\textquotesingle{s} focus is not limited to a specific language, our obtained evaluation results are highly competitive in an analogy among English state of the art methods.
\end{abstract}
\maketitle
\section{Introduction}
At the outset of its advent, fewer researches have successfully addressed style transfer\textquotesingle{s} applicability in Natural Language Processing (NLP) than Computer Vision \citep{gatys2016preserving, luan2017deep, gatys2015neural}. Besides having no reliable evaluation metric, parallel corpora shortage \citep{fu2017style} was another impediment to slow down its advance in natural language applications. But as robust pre-trained language models took the lead in natural language generation tasks and both manual and automatic evaluation metrics appeared, challenges of applying style transfer for text data got gradually overcame. Such developments eventuated in considerable growth of Text Style Transfer\textquotesingle{s} significance among other NLP tasks. However, the monolingual approaches of previous studies can be recalled as their primary pitfall.  \\
\-\hspace{0.40cm} Notwithstanding the shortage of resources, Persian NLP has recently witnessed a plethora of advancements, albeit to the best of our knowledge, no one has experimented with a Text Style Transfer method in Persian. Lack of corpora, ambiguous semantics, and exacting pragmatic are among the most substantial challenges of processing this natural language. \\
\-\hspace{0.40cm} In this work, we introduce the foremost instance of a multilingual style transfer method called \textbf{PGST}, a \textbf{P}olyglot \textbf{G}ender \textbf{S}tyle \textbf{T}ransfer method, which mainly revolves around transferring the style of a sentence by the gender of its author. The proposed style transfer method in this research relies on the power of a character-based token classifier, a pre-trained word embedding, and an exclusive Beam Search decoder, which each make up different building blocks of our method. Having trained the first and employed the last two, the task of Text Style Transfer applies to different natural languages provided the prerequisites mentioned earlier. But on the debit side, Style Transfer has just recently reached out to text data, making it a laborious and very time-consuming task to evaluate \citep{mir2019evaluating}. Besides, assessing transferred outputs to a multilingual extent demands a mutual set of evaluation metrics that assess different linguistic characteristics in common terms. Hence we evaluated our approach using automatic, statistical, and human-judgment-based scores to highlight its success, which will be discussed in great detail in the evaluation section of this paper.
\\
\-\hspace{0.40cm} To transfer the style of a text from one gender to another; first, we must understand their essential differences. Based on sociolinguistic studies, men and women have been shown to have distinct, deeply rooted variations in their language \citep{WALLENTIN202081}. Such linguistic phenomena may have been caused due to non-identical social and psychological circumstances that men and women undergo throughout their lives \citep{Jinyu2014StudyOG}. Such studies paved the way for us to address style transfer with far more excellent knowledge and base our approach on such differences, which emerge mostly in specific parts of speech tags. On the other hand, \citep{li-etal-2018-delete} simply overcame the task of attribute transfer by deleting identified attribute markers of text and replacing them with their equivalent retrieved target attributes. This work highly motivated us to distinguish gender-dependent words for gender style transfer purposes.
\\
\-\hspace{0.40cm} When transferring an input from its source style to to its transferred style  ($S_t$ ), preserving an input\textquotesingle{s} content and fluency are among the two most critical challenges that a style transfer method faces. In our introduced method, given an input document, content\textquotesingle{s} resistance to change is being handled by suggesting replacements from each token\textquotesingle{s} embedding space. The fluency aspect is under control by running a beam search algorithm on all suggestions, ranking suggestions continuously by our proposed scorer function. By following the preceding approach, we will have fluent transferred sentences with the same contextual information.\\
\-\hspace{0.40cm} The contributions of this paper are as follows:
\begin{itemize}
\item We propose a new Text Style Transfer method, able to be applied on a wide range of natural languages. Founded on sociolinguistics knowledge and through utilizing neural networks, pretrained word embeddings, and the beam search algorithm, we mainly focus on textual differences of men and women.
\item To evaluate our method in English and Persian, we proposed a multi-channel gender identification model, which achieves state-of-the-art result on our Persian dataset.
\item We have made all implementations and models of this paper publicly available at its GitHub Repository \footnote{https://github.com/Ledengary/PGST}.
\end{itemize}
\-\hspace{0.40cm} The remainder of this paper is structured as follows. In section \ref{rww}, we take a look at some related work, previously done in the scope of gender difference, text classification and text style transfer. Section \ref{ourapproach} is dedicated to giving an account of our proposed method. In section \ref{experiments} and \ref{discussion}, we share our experiments and discuss our method's outcome and finally, we conclude our paper in section \ref{conclusion}.

\section{Related Work}
\label{rww}
Our proposed method's central concept is structured on gender differences in written text, a well-studied sub-field of sociolinguistics. \cite{tru} and \cite{eck} carried out some preliminary work by focusing on male and female text's lexical and phonological differences. About a decade later, \cite{Bucholtz} laid the groundwork for the term "Sex Differences" and how the study of gender variations evolves based on theories as a complex and context-specific system. In the following decades, the field witnessed an overabundant domain-specific growth of gender-driven language studies in education, social networks, and science \citep{articlenahavandi, articlezare, LI2007301, METIN20112728, Cavas2010ASO, olivaaaa}. For instance, \cite{articleMusta} studied the language learning strategies of male and female students in five different categories (memory, compensation, cognitive, metacognitive, and social strategy). In comparison, the latest corpus-driven studies have scrutinized gender difference in word-level. They have called attention to the role of part-of-speech tags in gender language \citep{Pearce1, vbn, nbv, gdd}, specifically by studying the aftermath of different adjective choices on nouns. We drew inspiration from these long-studied sociolinguistics-based perspectives and led word-level gender differences to underlie our proposed Text Style Transfer method. \\
\-\hspace{0.40cm} Among text classification models introduced in Persian, the topic-model approach \citep{7585495} overcame the problems of dealing with Bag of Words, which considered each token as a feature, thus dealing with a vast number of elements and features inside a document. Besides, \cite{moradib} narrowed the task of text classification down to gender domain, where different statistical models such as Na\"ive Bayes, Alternating decision tree and Support vector machine were evaluated. Whereas a very small number of researches have successfully addressed author gender identification in Persian, a large number of previous studies have met this challenge in English under various terms \citep{company, bsir, CHENG201178, fatima2018, Martinc2018ReusableWF, martinc2019, istan, sotelo}. We believe our developed gender identification model, introduced as our automatic evaluator, takes a step toward further task advancements in Persian. \\
\-\hspace{0.40cm} In terms of Style Transfer, the English language has recently experienced rapid growth where several approaches are introduced using the latest architectures and algorithms. Having disentangled image features such as colour \citep{chen2016infogan}, \cite{hu2017controlled} focused on controlled text generation by learning disentangle latent representations and made a relation between an input\textquotesingle{s} style and content. Such models are trained hardly in an adversarial manner, which generates poor quality sentences as a result. Following Hu\textquotesingle{s} disentangling latent representation method, \citep{john-etal-2019-disentangled} recently proposed a simple yet efficient approach to approximate content information of bag-of-words features. Other researches, either directly or indirectly, have played several different parts in contributing to text style transfer. In particular, several models are introduced based on attention weights \citep{inproceedings}, neural machine translation \citep{s2018multipleattribute}, and deep reinforcement learning \citep{gong2019reinforcement}. Nevertheless, they came up short by misunderstanding input content, sparsity controversy and low-quality output, respectively. However, besides their flaws, these three approaches had a chief facet of contribution in common: using some of the most important deep learning algorithms in their proposed methods. Furthermore, leveraging pre-trained language models as discriminators \citep{yang2018unsupervised} in Generative-Adversarial-Network-based systems was another instance of how practical can an approach be, provided that it is built upon a pre-trained language model. However, this unsupervised model overcame the problem of the discriminator\textquotesingle{s} unstable error signal and should not be taken for granted. The same scenario happened in the Generative Style Transformer (GST) model \citep{sudhakar2019transforming}. It filled the quality loss gap caused by its Delete, Retrieve, Generate (DRG) framework \citep{li-etal-2018-delete} by powering up with a language model that made outputs\textquotesingle{} quality loss no more a debilitating concern. But besides all overcame dilemmas, limited available target-style data was still an issue where the domain adaptive model \citep{li2019domain} put an effort in solving the issue simultaneously as they kept an eye on relevant characteristics and content on the target domain in their approach.\\
\-\hspace{0.40cm} Akin to our domain-specific Text Style Transfer approach, where we concentrated primarily on the gender domain, recent task developments have metamorphosed into domain-specific methodologies as well. As a case in point, \cite{rao-tetreault-2018-dear} created a large corpus for benchmarking Style Transfer approaches and depicted Machine Translation's strength as a strong baseline, specifically in sentiment transfer. Moreover, based on Back-translation and Sentiment Analysis, \cite{inbook1112} introduced SentiInc to facilitate the task of sentiment-to-sentiment transfer by integrating sentiment-specific loss. Needless to mention that context plays a pivotal role in such tasks and has per se opened the doors of debate. Besides introducing two new datasets (Enron-Context and Reddit-Context), \cite{cheng2020contextual} developed CAST, a Context-Aware Style Transfer model, in which they allowed for the context being jointly considered alongside the style translation process by designing a Sentence Encoder and a Context Encoder in both presence and absence of parallel data settings. Based on their Sequence-to-sequence architecture \citep{sutskever2014sequence}, they strengthened both content preservation and context coherence and effectively took the first steps in modelling contextual details in Text Style Transfer. Additionally, \cite{zhou2020exploring} utilized a neural style component with an attention-based Sequence-to-sequence model to measure contextual word-level style relevance in an unsupervised setting. \\
\-\hspace{0.40cm} Regardless of all previously mentioned advancements, evaluating Text Style Transfer methodologies has been entangled \citep{mir2019evaluating}. Hence, to overcome this challenge, different evaluation metrics have been presented. Given $x$ and $x\textprime$ as the original and transferred text, \cite{fu2017style} introduced the following metrics which correlate considerably with human judgements:
\begin{enumerate}
    \item \textbf{Transfer Strength:} Motivated by \cite{shen2017style}, they used a classifier based on Keras examples\footnote{https://github.com/fchollet/keras/blob/
master/examples/imdb lstm.py}, which measures transfer accuracy.
    \item \textbf{Content Preservation:} To evaluate the similarity between $x$ and $x\textprime$, they calculated the cosine similarity of their relative embeddings.
\end{enumerate}
Similarly, \cite{mir2019evaluating} introduced the following:
\begin{enumerate}
    \item \textbf{Style Transfer Intensity:} Having mapped $x$ and $x\textprime$ to their style distributions, it quantifies the style difference of their distributions to alleviate evaluation.
    \item \textbf{Content Preservation:} Unlike \cite{fu2017style}, they utilize BLEU \citep{bleupaper} to measure the similarity between $x$ and $x\textprime$.
    \item \textbf{naturalness:} having passed $x\textprime$ to its function, it quantifies to what extent the transferred text is human-like.
\end{enumerate}
\cite{hu2020text} used the Perplexity Score ($PPL$) and Style Transfer Accuracy ($ACC$) to measure the fluency and transfer strength of $x\textprime$. Besides, Word Overlap ($WO$), Cosine Similarity, and self-BLEU were employed to measure content preservation. Finally, they summed up all metrics in the following two:
\begin{enumerate}
    \item \textbf{Geometric Mean ($G-Score$):} Which is equal to the mean of $1/PPL$, $WO$, $ACC$, and self-BLEU. (Note that they calculated the inverse of PPL since lower PPL is preferred and that the cosine similarity metric is not included in the mean due to its insensitivity)
    \item \textbf{Harmonic Mean ($H-Score$):} The Harmonic Mean of the above sub-metrics is calculated to highlight different priorities when evaluating.
\end{enumerate}
\-\hspace{0.40cm} We employed cosine similarity as one of our method\textquotesingle{s} crucial components to handle Content Preservation in this work. Additionally, We utilized ACC, BLEU, PPL, and a metric similar to Naturalness in our automatic and human judgments as well. Specifics of our evaluation metrics are broken down in detail in section \ref{experiments}.
\begin{figure*}[htb]
    \centering
    \includegraphics[width=.8\linewidth]{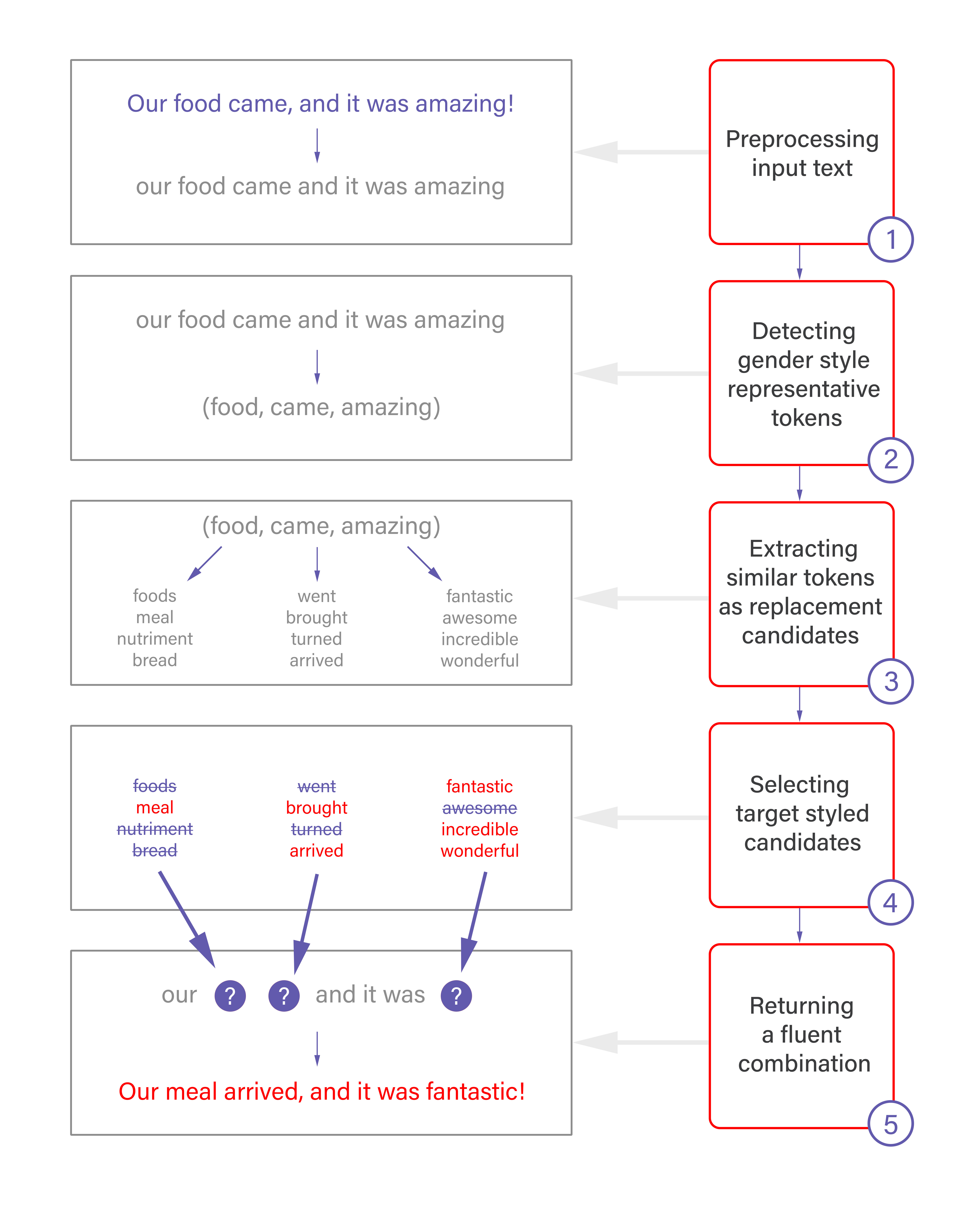}
    \caption{Perception of our proposed method\textquotesingle{s} different stages. text in blue/red are styled as male/female.}
    \label{fig:stages}
\end{figure*}
\begin{figure}[htb]
  \includegraphics[width=0.9\linewidth]{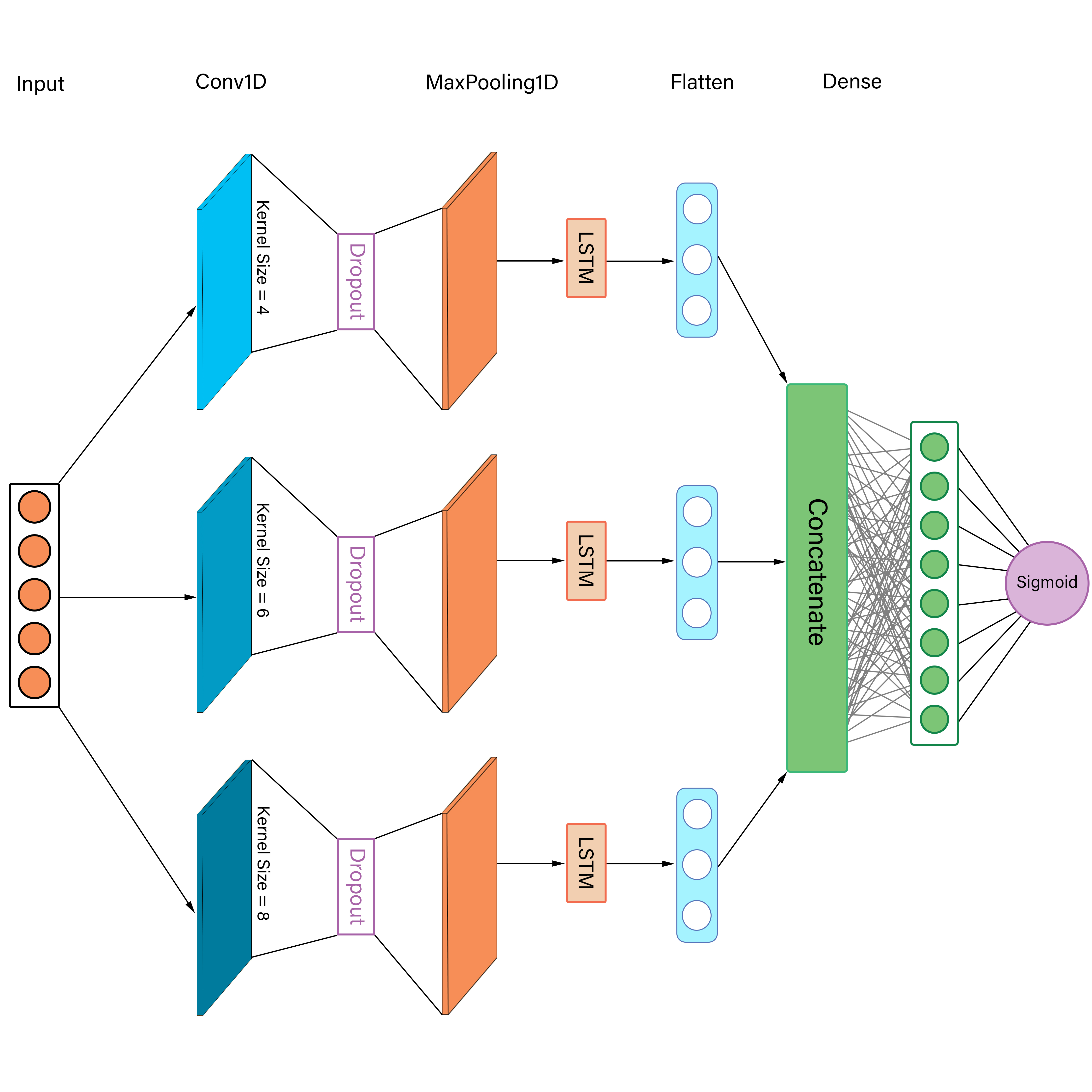}
    \vspace{\baselineskip}
  \caption{our baseline Gender Classifier\textquotesingle{s} neural network architecture.}
  \label{fig:CNNModelNew}
\end{figure}
\section{Our Approach}
\label{ourapproach}
Our proposed method is illustrated in Figure \ref{fig:stages}. We begin with pre-processing input documents by removing punctuation and stop words within documents (section \ref{bscS}). Then by detecting those specific tokens that make up the stylistic texture of an input document in section \ref{stS}, we have a set of tokens that we are willing to exchange with their appropriate opposite gender equivalents. In section \ref{bv3}, we use a pre-trained word embedding to obtain a similar set of tokens for each predetermined gender style representative token and a Character-based Token Classifier in section \ref{tccS} to select/eliminate all target/source styled candidates among suggested replacements. Finally, we pass all target styled tokens to our proposed Beam Search algorithm (section \ref{bmmcS}) to extract a verifiable combination of those tokens in terms of fluency. The details of the proposed method are discussed in more details in the following subsections.
\subsection{Baseline Gender Classifier}
\label{bscS}
Convolutional Neural Networks (CNN) have performed considerably better in terms of training time than other networks by peaking a better validation accuracy for small datasets with more consistency. We considered a 3-channel CNN layer architecture \citep{brown} with Long short-term memory (LSTM) layers on top for our specified gender classification task. The purpose of using a multi-channel CNN architecture is to proceed with documents in different resolutions (or n-grams) at each time step by defining different kernel sizes for each channel\textquotesingle{s} convolutional layer. Although CNNs are generally used in computer vision, they have performed exceptionally well in capturing input text patterns. The necessity of LSTM layers\textquotesingle{} presence in our architecture is that the model needs to memorize these extracted patterns. LSTM layers have an exclusive internal mechanism which is composed of \textit{forget}, \textit{input} and \textit{output} gates with a \textit{state} cell, in order to regulate the flow of given patterns (Equations \ref{eq:e1}-\ref{eq:e6}). Therefore, by locating LSTM layers on top of Convolutional layers, LSTM fulfils such demand. (model visualized in Figure \ref{fig:CNNModelNew}).
\begin{equation}
f_t = \sigma_g(W_fx_t + U_fh_{t-1} + b_f)
\label{eq:e1}
\end{equation}
\begin{equation}
i_t = \sigma_g(W_ix_t + U_ih_{t-1} + b_i)
\label{eq:e2}
\end{equation}
\begin{equation}
o_t = \sigma_g(W_ox_t + U_oh_{t-1} + b_i)
\label{eq:e3}
\end{equation}
\begin{equation}
\tilde c_t = \sigma_g(W_cx_t + U_ch_{t-1} + b_i)
\label{eq:e4}
\end{equation}
\begin{equation}
c_t = f_t \circ c_{t-1} + i_t \circ \tilde c_t
\label{eq:e5}
\end{equation}
\begin{equation}
h_t = o_t \circ \sigma _h(c_t)
\label{eq:e6}
\end{equation}
\-\hspace{0.40cm} As shown above, we have $f_t$, $i_t$ and $o_t$ each notated as an LSTM layer\textquotesingle{s} forget, input and output gates\textquotesingle{} activation vectors and $h_t$ as the layer\textquotesingle{s} final output. Henceforth by calling a document\textquotesingle{s} source style as $S_s$ and its target style as $S_t$, we begin to transfer a document\textquotesingle{s} style from $S_s$ to $S_t$ in upcoming sections of \ref{stS} to \ref{bmmcS}.   

\subsection{Detecting Gender Style Representative Tokens}
\label{stS}
In a considerable number of previous studies, some specific part-of-speech tags have reminisced as substantive pivotal points in determining the author\textquotesingle{} gender. In most languages, and more specifically in our Persian corpus, depending  on its label, each gender tagged document has a set of words that play a document\textquotesingle{s} most significant stylish roles, which are mostly categorized as either Adjective or Adverb. The margin that differentiates the two genders is mostly made by the author\textquotesingle{s} choice of words in these specific part-of-speech tags. Adjectives, as the most determinative part-of-speech tag, is composed of different types which differ based on the language. The following list contains English and Persian types of adjectives.
\begin{itemize}
    \item \textbf{English:} Adjectives are grouped either as \textbf{Descriptive} or \textbf{Limiting}. Whereas the former describes the quality of the noun, the latter limits it. Descriptive adjectives bifurcate in terms of where they are located. If the adjective appears directly beside the noun it describes, it is called an Attributive Adjective (e.g. \textit{The restaurant has a \textbf{remarkable} view!}) and if connected to the noun with a linking verb, they are called Predicate Adjectives (e.g. \textit{The pizza was too \textbf{salty}!}). Besides Definite \& Indefinite Articles, Limiting adjectives are categorized as 8 different subgroups of Possessive, Demonstrative, Indefinite, Interrogative, Cardinal, Ordinal, Proper, and Nouns used as adjectives. \\
    \item \textbf{Persian:} Similar to English and most other languages, Persian adjectives also consist of Descriptive Adjectives. The only exception is that unlike English, Attributive Adjectives come after the noun. Other Persian Adjective types are Cardinal, Ordinal, Interrogative, Indefinite, and Exclamatory Adjectives.
\end{itemize}
\-\hspace{0.40cm} Since we are studying gender differences to a multilingual extent, we mainly focus on Descriptive Adjectives, which is common among almost all natural languages. Hence, to further break down other types of adjectives is out of this paper's scope, and we leave it to future work to come up with a more unified set of mutual adjectives among multiple languages. \\
\-\hspace{0.40cm}To detect the specified tags, we used a part-of-speech tagger that would tag each token of a document to search for an elegant replacement token from its opposite gender in the subsequent step. We used Parsivar \citep{mohtaj-etal-2018-parsivar} and Spacy \citep{honnibal-johnson-2015-improved} language processing toolkits\textquotesingle{} part-of-speech taggers to do so on our Persian and English corpora, respectively.
\subsection{Extracting Similar Tokens as Replacement Candidates}
\label{bv3}
By detecting gender style representatives of an input text, we look after replacements from which we may bear down on style transfer purposes. We used fastText \citep{arm2016bag} word vectors as a pre-trained word embedding that was trained using Continuous Bag of Words (CBOW) with character n-grams of length 5, a window of size 5 and 300 in dimension, specified to return the $top_n$ most similar words of a given token. \\
\begin{equation}
\label{eq:simeq}
similarity = \cos (\theta)= {{\bf A} \cdot {\bf B} \over \|{\bf A}\| \|{\bf B}\|} = \frac{ \sum_{i=1}^{n}{{\bf A}_i{\bf B}_i} }{ \sqrt{\sum_{i=1}^{n}{{\bf A}_i^2}} \sqrt{\sum_{i=1}^{n}{{\bf B}_i^2}} }
\end{equation}\\
\-\hspace{0.40cm} Given \textbf{A} and \textbf{B} as two predetermined words' word vectors and $\textbf{A}_i$ and $\textbf{B}_i$ as their vector components, the cosine of the angle between word vectors is calculated using Equation \ref{eq:simeq}. By considering $A$ as a gender style representative's word vector and $B$ as for all other available word vectors in the fastText\textquotesingle{s} vocabulary, we apply this computation on all $(A, B)$ pairs. Altogether, whenever a token is categorized as either Adjective or Adverb (or any other specified tag). In that case, we pass the token to the $most\_similar$ built-in function of fastText to obtain a set of suggested replacement tokens with their specific similarity rates to choose between. However, to transfer a document\textquotesingle{s} style, we have to select replacement tokens from the opposite gender suggested ones. This is where our character-based token classifier indicates from which gender are the suggested tokens derived.
\begin{figure}[htb]
    \centering
    \includegraphics[width=0.8\textwidth]{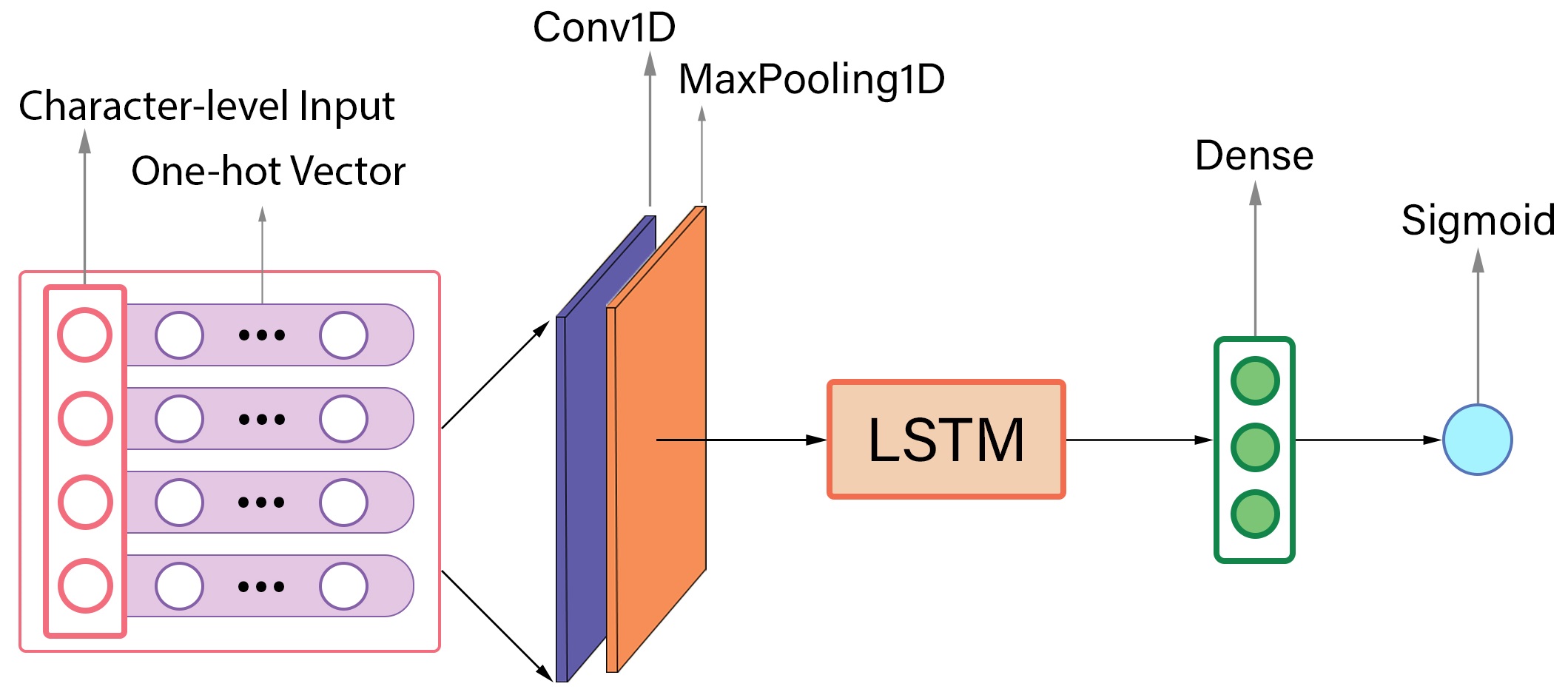}
    \vspace{\baselineskip}
    \caption{our Character-based Token Classifier\textquotesingle{s} neural network architecture.}
    \label{fig:charModel}
\end{figure}
\subsection{Selecting Target Styled Candidates}
\label{tccS}
\begin{figure}[htb]
  \centering
  \includegraphics[scale=0.6]{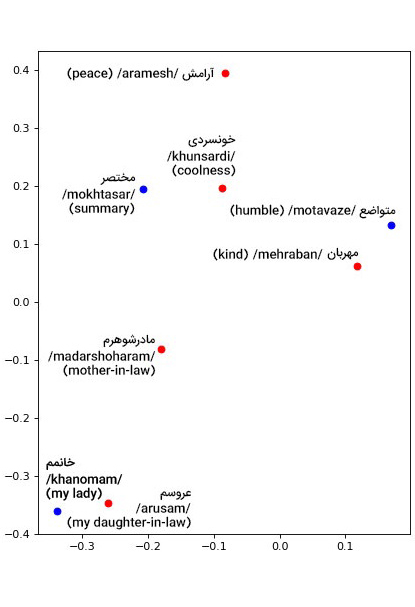}
    \vspace{2\baselineskip}
  \caption{An example of fastText Word Embedding Space that has been projected with PCA. Note: each Blue/Red Scatter represents Male/Female classified. (Note: Persian pronunciations are shown between slashes and english translations are included between parenthesis.}\label{fig:my_label1}
\end{figure}
Besides classifying documents, we need to train a new model to classify tokens as either male or female. This opportunity allows us to waive those suggested replacement tokens with the same style as $S_s$, and leave the set only with the $S_t$ styled ones. We used a sequential neural network with a Convolutional Layer and a Long Short-Term Memory (LSTM) layer on top. We considered a character-based representation of a token with one-hot vector representations specified for each character of an input token. A visualization of this model is shown in Figure \ref{fig:charModel}. \\
\-\hspace{0.40cm}The reason behind using a character-based model when classifying a single token is to handle the unfortunate probability that a token is out of embedding\textquotesingle{s} vocabulary (OOV) \citep{mam} or is misspelt \citep{brown}. In either way, if the model is character-based, it would automatically find the best pattern to digest the token and represent it as a vector. Last but foremost, character-based models have performed superior to word-based models when it comes to capturing hidden emotions from an input \citep{zhang2015characterlevel}. Furthermore, using word-based models for token classification was much more likely to get biased towards a specific output label. We feed the model with all extracted Adjectives, Adverbs, Verbs, and Nouns from both male/female documents so the model would categorize the word embedding\textquotesingle{s} suggested tokens as either male or female. A visualized result of applying this model on a word embedding space is shown in Figure \ref{fig:my_label1}. \\
\-\hspace{0.40cm} It is worthy to mention that extra filters are applied on specific part-of-speech tags\textquotesingle{} suggested lists. For instance, we step through each replacement candidate of a verb, and remove those with the same stem or lemma. This allows us to only focus on candidates with different origins. Additionally, we transform each candidate to comply with both tense and form of the original token for which it is suggested as replacement. Hence, the method avoids replacing tokens with candidates that will damage the flow and fluency of the original text.
\begin{figure}[htb]
    \centering
    \includegraphics[width=\textwidth]{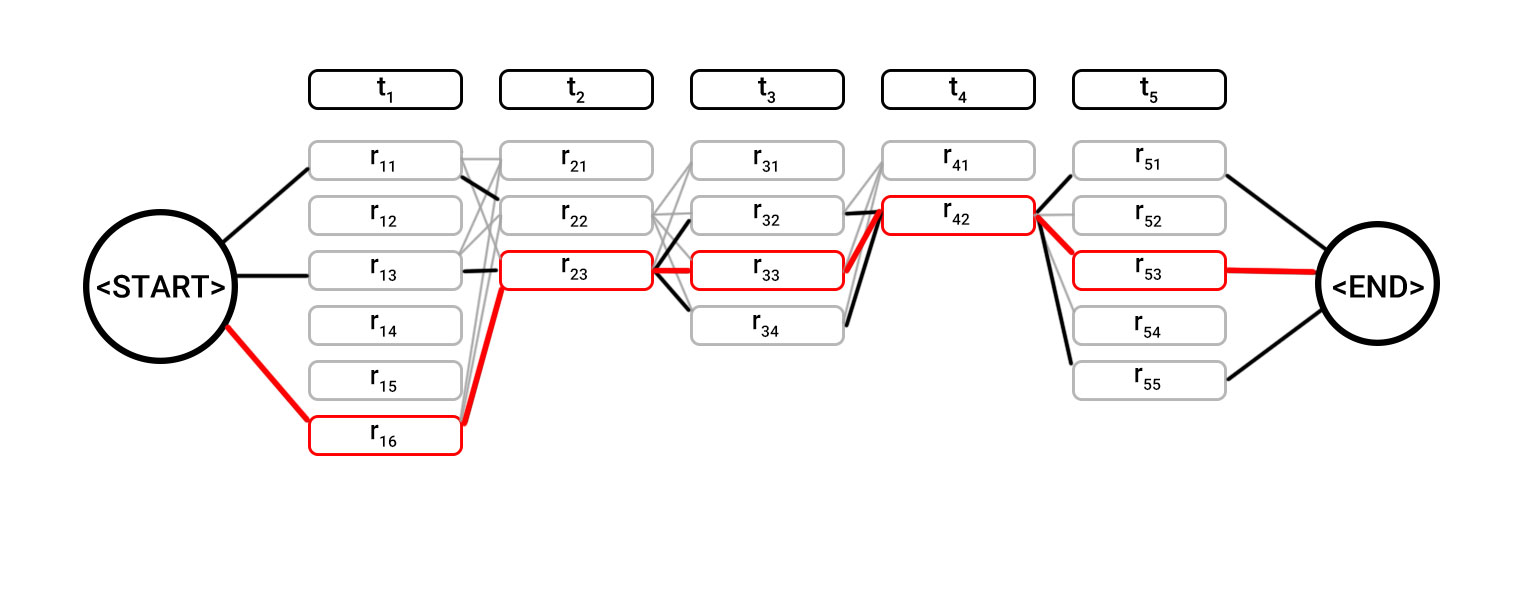}
    \vspace{2\baselineskip}
    \caption[width=\textwidth]{An overview of what our model\textquotesingle{s} approach for transferring an input\textquotesingle{s} style from $S_s$ to $S_t$ does. Note: for an input with five tokens, we have  $t_{1..5}$  and each  $t_i$  has a set of $r_{ij}$ with j as the number of opposite-gender predicted set of the Token Classifier between the $top_n$=10 Word Embedding\textquotesingle{s} most similar suggested words.}
    \label{fig:my_label2}
\end{figure}
\subsection{Returning a Fluent Combination}
\label{bmmcS}
By now, we have a set of target styled replacement tokens for each word in the document, and our only concern is to choose the most probable combination in terms of fluency. Since the processing in document-level costs dealing with much higher probable token combinations than sentence-level, extracting the desired token combination requires an optimized algorithm and heuristic. The designated algorithm in most neural machine translation systems is Beam Search. This heuristic searching algorithm expands the K (beam width) most probable children of a given node and only keeps track of the K most probable traversed paths. \\
\-\hspace{0.40cm} To keep up the fluency of a given document, we keep track of all unigram, bigram, trigram, and 4-gram counts in our Baseline Gender classifier\textquotesingle{s} train set. To do so, we define a dictionary and iterate over all documents and extract their specified n-gram counts and assign them as keys and their counts as values. This dictionary will further be used in the scorer function below:
\begin{equation}
    BeamScore = \frac{4 \times f + 3 \times t + 2 \times b + 1 \times u}{4 \times 10 \times (1 - sim)}
\label{eq:e7}
\end{equation}
\begin{algorithm}[htb]
\SetAlgoLined
\textbf{Data:} suggestions, BW \\
\KwResult{BW most probable decoded texts}
 Add $\{<END>\}$ to $suggestions$\;
 $beams = \{\{<START>\}\}$\;
 $scores(\{<START>\}) = 0$\;
 \For{$row \in suggestions$}{
    $candidates = \{\}$\;
    \For{$b \in beams$}{
        \For{$t \in row$}{
            $c\textprime = b + \{t\}$\;
            $s = BeamScore(b.top(1), b.top(2), b.top(3), t)$\;
            Add $c\textprime$ to $candidates$\;
            $scores(b) += s$\;
        }
    }
    \small $beams = bestBeams(candidates,BW)$\;
 }
 \caption{\label{alg:beam}Our Proposed Beam Search}
\end{algorithm}
\-\hspace{0.40cm} Given $f$, $t$, $b$, $u$, and $sim$ as the 4-gram, trigram, bigram, unigram, and the similarity rate (calculated by word embedding for each of the suggested tokens), we calculate BeamScore (equation \ref{eq:e7}) which is the mean of standardized n-gram counts and divides it by the dissimilarity of the tokens. We designate the efficacy of 40\%, 30\%, 20\% and 10\% to each of the 4 to 1-gram counts, respectively. However, the dilemma here is to score the first and the last word of a sentence and examine a suggested token's score to start or end a sentence with. Therefore, the two tags of <START> and <END> are added as tokens to each document\textquotesingle{s} beginning and end to overcome the problem. Having defined the root as <START> and replacement tokens as nodes, the algorithm calculates at each step the beam score based on the traversed path's last three nodes and the visiting node until it reaches the <END> node. By the end, the algorithm returns a fluent combination of tokens since we extracted the most probable sentence and transferred an input document from $S_s$ to $S_t$ meanwhile. The pseudocode of our implemented beam search is given in Algorithm \ref{alg:beam}. \\
\-\hspace{0.40cm} By transferring the test set and passing it again to our gender classifier model, we measure our model\textquotesingle{s} accuracy loss on $S_t$ text, which it formerly predicted in their $S_s$ style. An overview of our style transfer approach is visualized in Figure \ref{fig:my_label2}.\\
\section{Experiments}
\label{experiments}
We kick this section off by first breaking down our employed datasets and previously mentioned models by their hyperparameter choices. We then demonstrated our achieved results in great detail and performed human, statistical, and automatic evaluations to obtain a full-scale understanding of our approach\textquotesingle{s} functionality.
\subsection{Experimental Setup}
\setcounter{table}{-1}
\begin{table}[htb]
\centering
  \tbl{\caption{Dataset Comparison}}
 {
    \begin{tabular}{@{\extracolsep{\fill}}lcrrrr}
    \hline
    Dataset & Style & Train & Dev & Test & Overall \\
    \hline
    \multirow{2}{*}{{ Persian}} & { Male} & { 5,120} & { 1,329} & { 1,591} & { 8,040}\\ \cdashline{2-6}
    & { Female} & { 8,458} & { 2,066} & { 2,653} & { 13,177}\\
    \hdashline
    \multirow{2}{*}{{English}} & { Male} & { 2,062,289} & { 257,787} & {\ 257,786} & { 1,288,931}\\ \cdashline{2-6}
    & { Female} & { 2,062,289} & {257,787} & { 257,786} & { 1,288,931}\\
    \hline
    \end{tabular}}
  {\begin{tabnote}
  \end{tabnote}}
  \label{tab:mainDataset}
\end{table}
\textbf{Dataset:} The following two datasets of Formal Gender Tagged Persian Corpus \citep{moradib} and Gender (English) by \citep{reddy}, are used for our experiments. The Persian dataset contains 245 different documents for each of the two male/female gender labels, including documents mostly books or stories by either a male/female author based on its label. Since each document\textquotesingle{s} size varies in a broad span of tokens, each document is broken into subdocuments with smaller lengths. On the contrary, the gender tagged dataset is built-up by sentence-level reviews of different food businesses on Yelp, each classified as male/female. A comparison between the two datasets is shown in Table \hyperref[tab:mainDataset]{1}.\\
\textbf{Hyperparameters:} There is an Embedding, a Convolutional and an LSTM layer prepared in each of our Baseline gender classifier\textquotesingle{s} channel with the output dimension of 100 for Embedding layers, 32 filters, a dropout rate of 0.5 and a max pool size of 2 for Convolutional layers and 256 hidden units with a recurrent dropout rate of 0.2 for LSTM layers. Each channel\textquotesingle{s} Convolutional layer has a different value of 4, 6, and 8 assigned to its kernel Size to process documents at different n-grams. The model is trained for 10 epochs with these hyperparameter choices. \\
\-\hspace{0.40cm} In our character-based token classifier, we sequenced the same setup in a single channel as the previous model, a Convolutional and LSTM layer is stacked up, with 32 filters, 8 in kernel size, max pool size of 2 for the Convolutional layer and 125 hidden units for the LSTM layer. This model was trained for the same number of epochs as our baseline model with no dropouts. \\
\-\hspace{0.40cm} In both models, Adam \citep{kingma2014adam} was recruited as the optimizer with its learning rate set to 0.001 since it performs better in handling sparse gradients.
\setcounter{table}{0}
\begin{table}[h!]
\centering
  \tbl{\caption{Model Comparison}}
 {
    \begin{tabular}{@{\extracolsep{\fill}}lcc}
    \hline
    \multirow{2}{*}{Model} &
      \multicolumn{2}{c}{Accuracy} \\ \cline{2-3}
    & Persian & English \\
    \hline
     Na\"ive Bayes & 69\% & 73\% \\ \hdashline
     Logistic Regression & 62\% & 70\%\\ \hdashline
     Multi-lingual BERT & 65\% & 75\%\\ \hdashline
     SVM \citep{moradib} & 72\% & -\\ \hdashline
    \textbf{CNN + LSTM NN} & \textbf{90\%} & \textbf{81\%}\\ 
    \hline
    \end{tabular}}
  {\begin{tabnote}
  \end{tabnote}}
  \label{tab:dataComp}
\end{table}
\subsection{Model Evaluation} 
\label{sec:modeleval}
To acquire the highest accuracy possible, we have stepped through different models and architectures. The process of choosing our gender classification model was based on a comparison between Na\"ive Bayes, Logistic Regression, Multi-lingual BERT, and CNN + LSTM (Baseline Classifier) Neural Network architecture. As shown in Table \hyperref[tab:dataComp]{2}, our proposed Baseline Classifier architecture peaked the highest accuracy for our classification problem on Persian text with 90\% and 80\% in English. We finalize our model here since an efficient model in both languages is our main concern. \\
\-\hspace{0.40cm}Due to the relatively small amount of data in our Persian corpora, a probabilistic model like Na\"ive Bayes performs poorly with very low precision and recall, as long as the frequency-based probability estimate becomes zero for a value with no occurrences of a class label. \\
\-\hspace{0.40cm} Pre-trained language models like Bi-directional Encoder Representations from Transformer \citep{devlin2018bert} or BERT have had a significant rise due to their success in topping state-of-the-art natural language processing tasks. However, at the time of our research, there has not been any Persian specific pre-trained transformer language model introduced. But between proposed pre-trained models of BERT, multilingual cased contains the top 100 languages with the largest Wikipedias including Persian (Farsi), which, as shown, fine-tuning it did not perform as durable as training a classifier from the scratch. \\
\-\hspace{0.40cm} A logistic regression model is a generalized linear model that could be reminded of a neural network with no hidden layers. Evidently, a neural network model with such hidden layers as Convolutional and LSTM carries more advantages in solving our problem. Convolutional layers perform outstandingly in pointing out tokens that are good indicators of an input\textquotesingle{s} class, and LSTM layers in associating both short and long-term memory with the model, thus resulting in a better accuracy score in an analogy to Support vector machine (SVM) algorithm \citep{moradib}.\\
\-\hspace{0.40cm} When designing a token classifier, casting each token to its 300-dimensional embedding space representation as model inputs results in contextual information loss, making a word-based model an inappropriate choice for classifying tokens as male/female. On the other side, designing a model on a character-level fills the gap and distributes tokens by stylish criteria in a better way. \\
\setcounter{table}{1}
\begin{table}[htb]
\centering
  \tbl{\caption{Results of defeating Gender Identification with different style transfer approaches and models}}
 {
    \begin{tabular}{@{\extracolsep{\fill}}lcccccc}
    \hline
    \multirow{2}{*}{\qquad Approach} &
      \multicolumn{3}{c}{Persian Accuracy} &
      \multicolumn{3}{c}{English Accuracy} \\  \cline{2-7}
    & $T_a$ & $T_r$ & $T_w$ & $T_a$ & $T_r$ & $T_w$ \\
    \hline
     1 \-\hspace{0.20cm} Word-based + (Adj, Adv) & 86\% & 92\% & 15\% & 77\%	& 95\% & 16\% \\ \hdashline
    2 \-\hspace{0.20cm} Word-based + (Adj, Adv, V) & 83\% & 90\% & 20\% & 70\% & 88\% & 22\% \\ \hdashline
    3 \-\hspace{0.20cm} Word-based + (Adj, Adv, V, N) & 69\% & 62\% & 33\% & 68\% & 79\% & 31\% \\ \hdashline
    4 \-\hspace{0.20cm} \textbf{Character-based + (Adj, Adv, V, N)} & \textbf{65\%} & \textbf{56\%} & \textbf{36\%} & \textbf{34\%} & \textbf{61\%} & \textbf{37\%} \\
    \hline
    \end{tabular}}
  {\begin{tabnote}
  \end{tabnote}}
  \label{tab:mainComp1}
\end{table}
\-\hspace{0.40cm} Achieved experimental results by repassing the test set to our Gender Classifier depicts our paper\textquotesingle{s} success the most. It demonstrates how have different approaches resulted, using different models and architectures on our set. We call our gender classifier\textquotesingle{s} $S_s$ test set as $D_a$ and divide it into two different subsets: 1) $D_r$, which includes all documents that have been predicted correctly by the model, and 2) $D_w$, which consists of all documents that have been mispredicted by the model. Besides, by transferring all $D_a$ documents to $S_t$, we name the transferred set as $T_a$, $D_r$ as $T_r$ and $D_w$ as $T_w$. By taking a look at Table \hyperref[tab:mainComp1]{3}, it has been clearly demonstrated how the primary approach has elevated by changing its different components. Each approach\textquotesingle{s} name contains two vital information. a) what its token classifier model was based on (word or character) b) what tags are supposed to be identified by our part-of-speech tagger to be changed into their opposite gender. Our goal is to defeat the gender identification model by transferring a sentence\textquotesingle{s} style to its opposite gender, thus diminishing the gender identification model\textquotesingle{s} accuracy, meaning the model performs poorly in identifying inputs style. As shown in the table, the more robust our token classifier and the more varied our part-of-speech tags scope gets, the weaker the gender identification model\textquotesingle{s} performance gets. Each of the three $D_a$ , $D_r$ and $D_w$ sets have their own accuracy scores. Since $D_a$ is the test set on which the model performed 90\% in Persian and 80\% in English, $D_r$ and $D_w$ as subsets of $D_a$ include those the model predicted correctly and incorrectly, would evidently result in 100\% and 0\% accuracy by passing only the subsets themselves to the model for prediction purposes. As mentioned before, by transferring these raw sets, we have $T_a$, $T_r$ and $T_w$ from which we expect lower accuracies in an analogy to their source styled sets. The first approach resulted in a 4\% decrease overall, 8\% on faking the ones it had predicted once correctly, but unintentionally helping the model to predict the ones it had mistaken before correctly. Meaning the approach has helped the model instead of faking it. 
\setcounter{table}{2}
\begin{table}[htb]
\centering
  \tbl{\caption{A comparison of positive and negative effects of applying different approaches}}
 {
    \begin{tabular}{@{\extracolsep{\fill}}lrrrrrr}
    \hline
    \multirow{2}{*}{\qquad Approach} &
      \multicolumn{3}{c}{Persian} &
      \multicolumn{3}{c}{English} \\
      \cline{2-7}
     & {f} & {h} & {Trade-off} & {f} & {h} & {Trade-off} \\
    \hline
     1 \-\hspace{0.20cm} Word-based + (Adj, Adv) & 306 & 62 & 5.74 & 10,079 & 8,993 & 0.42 \\ \hdashline
    2 \-\hspace{0.20cm} Word-based + (Adj, Adv, V) & 382 & 83 & 7.04 & 24,189 & 12,366 & 4.58 \\ \hdashline
    3 \-\hspace{0.20cm} Word-based + (Adj, Adv, V, N) & 1,455 & 137 & 31.05 & 42,331 & 17,424 & 9.66 \\ \hdashline
    4 \-\hspace{0.20cm} \textbf{Character-based + (Adj, Adv, V, N)} & \textbf{1,684} & \textbf{150} & \textbf{36.14} & \textbf{122,963} & \textbf{20,797} & \textbf{39.63} \\
    \hline
    \end{tabular}}
  {\begin{tabnote}
  \end{tabnote}}
  \label{tab:tfv}
\end{table}
\setcounter{table}{3}
\begin{table}[htb]
\centering
  \tbl{\caption{A contingency table on our finalized style transfer approach (i.e., Character-based + (Adj, Adv, V, N)) in Persian and English}}
 {
    \begin{tabular}{@{\extracolsep{\fill}}lrrrr}
    \hline
    \multirow{2}{*}{\qquad\quad Set} &
      \multicolumn{2}{c}{Persian} &
      \multicolumn{2}{c}{English} \\
      \cline{2-5}
    & correctly & incorrectly & correctly & incorrectly \\
    \hline
     $D_a$ (all docs) & 3,828 & 416 & 201,579 & 56,207\\ \hdashline
     $D_a$ (male docs) & 1,460 & 151 & 102,592 & 25,976\\ \hdashline
     $D_a$ (female docs) & 2,368 & 265 & 98,987 & 30,231\\ \hdashline
     $T_a$ (all docs) & 2,758 & 1,486 & 87,647 & 170,139\\ \hdashline
     $T_a$ (male docs) & 1,040 & 571 & 66,133 & 62,435\\ \hdashline
     $T_a$ (female docs) & 1,718 & 915 & 21,514 & 107,704\\
     \hline
    \end{tabular}}
  {\begin{tabnote}
  \end{tabnote}}
  \label{tab:tb5}
\end{table}
\begin{equation}
Trade-off = \frac{f - h}{n} \times 100 
\label{eq:to}
\end{equation}
\-\hspace{0.40cm} In equation \ref{eq:to}, by naming the number of documents that faked the model $f$ and the number of documents that helped as $h$ and n as the total number of test set documents, we acquire a trade-off value that the higher it gets, the more effective its approach is. Persian $D_a$ contains 4,244 test documents in which 3,828 documents were correctly guessed and belong to set $D_r$ and 416 documents guessed incorrectly by the gender identification model, which belongs to set $D_w$. As shown in Table \hyperref[tab:tfv]{4} there was an 8\% decrease in $D_r$\textquotesingle{s} accuracy (251 documents) and a 15\% rise in $D_w$ (166 documents), resulting in a trade-off value of 2, which demonstrates its lack of ability in defeating gender identification. But as we go along testing approaches 2, 3 and 4, we get back the trade-off values of 2.2, 21.84, and 23.13. The major leap between second and third model\textquotesingle{s} trade-off values represents the pivotal role that a bigger scope of part-of-speech tags play and finally, decreasing the $T_w$ value and giving rise to $T_a$ and $T_r$ values by changing the token classifier\textquotesingle{s} base between approaches three and four. The same evaluations have been made in English as we get back the results of 34\%, 61\%, and 37\% for $T_a$, $T_r$ and $T_w$, and the trade-off value of 39.63. Specific contingencies of applying our finalized approach (i.e., Character-based + (Adj, Adv, V, N)) on our defined sets of $D_a$ and $T_a$ in both languages is shown in Table \hyperref[tab:tb5]{5}, which demonstrates the number of documents that were predicted either correctly or incorrectly by our baseline gender classifier. \\
\-\hspace{0.40cm} In order to prevent the transfer approach to unintentionally help the classifier to predict the documents correctly, amplifying our character-based token classifier is the most rational alternative, since converting a document\textquotesingle{s} content to its target style is its primary essence, and content is what the classifier is obligated to detect.
\subsection{Statistical Evaluation}
To determine whether there is a significant difference between the means of the two gender labels, we use a statistical hypothesis testing tool called T-Test \citep{ttestpaper} to assure that there is not an unknown variance and that labels are all distributed normally.\\
\-\hspace{0.40cm} In order to assure if the two gender labels come from the same population, by taking samples from each of the two labelled sets, T-Test hypothesises that the two means are equal. By calculating certain values and comparing them with the standard values afterwards, T-Test decides whether inputs are strong and not accidental or that they are weak and probably due to chance, resulting in rejection and acceptance of the hypothesis, respectively. \\
\begin{equation}
t = \frac{\overline{X}_D - \mu_0}{s_D / \sqrt{n}}
\label{eq:ttestt}
\end{equation} 
\-\hspace{0.40cm} We use inputs in both original and transferred style in equation \ref{eq:ttestt} to measure the t statistic value for dependent paired samples in which $\overline{X}_D$ and $s_D$ are each pair’s average and standard deviation of their difference, $n$ as the number of pairs, and $\mu_0$ as hypothesized mean, which we assign to zero when testing the average of the difference. \\
\-\hspace{0.40cm} The p-value is the probability of obtaining an equal or more extreme result than the one obtained when the hypothesis is true. $significance$ $level$ (or alpha) is a threshold value which is the eligible probability of making a wrong decision (rejecting the hypothesis). We have calculated P-values in both languages by considering n-1 degrees of freedom and assigning 0.01 to alpha. As demonstrated in Table \hyperref[tab:ttesttable]{6}, test results are significant in both languages with their acquired p-values.
\setcounter{table}{4}
\begin{table}[htb]
\centering
  \tbl{\caption{P-Values for paired samples in our corpora. (Alpha = 0.01, degree of freedom = n-1)}}
 {
    \begin{tabular}{@{\extracolsep{\fill}}lr}
    \hline
    \footnotesize Language & \footnotesize P-Values \\
    \hline
     \footnotesize Persian & \footnotesize 1.64813366054e-10 \\
     \footnotesize English & \footnotesize 4.04649941132e-3 \\
     \hline
    \end{tabular}}
  {\begin{tabnote}
  \end{tabnote}}
  \label{tab:ttesttable}
\end{table}

\subsection{Human Evaluation}
Generated text need to be assessed in order to prove their correctness. We considered facets like $fluency$ and $semantic$ that each sample had to be assessed based on. The former facet determines whether a given text is reasonably close to legible human language or that it is presented in an indecipherable manner. As the name implies, the latter is employed to evaluate inputs depend on conceptual meanings when interpreted. We additionally added an $adulteration$ facet to determine whether the given text seemed adulterated or not. Randomly, we sampled 300 inputs, including $S_s$ styled and 75 $S_t$ styled text in both English and Persian. We assigned each 100 samples to an annotation group with three different annotators by shuffling inputs and dividing them by three. We then asked annotators to rank each sample based on given criteria in binary representation. The reason behind using $S_s$ text from both English and Persian corpora is to uniform all inputs to figure if $S_t$ inputs seem evidently adulterated among the others or that they are formed in a logical and acceptable form.
\subsubsection{Inter-annotator agreement}
\-\hspace{0.40cm}Before particularizing annotations, we test interrater reliability with kappa \citep{kappapaper}, a standard measure of inter-annotator agreement (IAA) which aims to compare the amount of agreement that we are actually getting between judges to the amount of agreement that we would get purely by chance.\\
\-\hspace{0.40cm}By letting $N$ be the number of documents and defining $R$ and $I$ as two sets of agreed and disagreed documents for each of the three annotators in a specific group, $A$ would be the set of documents where all three annotators agreed on and $P(A)$ and $P(E)$ as fractions of real and accidental agreements.\\
\\
\\
\begin{equation}
A = (R_1 \cap R_2 \cap R_3) \cup (I_1 \cap I_2 \cap I_3)
\label{eq:kappaeq}
\end{equation} 
\begin{equation}
P(A) = \frac{|A|}{N}
\label{eq:kappaeq}
\end{equation} 
\begin{equation}
P(E) = (\frac{R_1}{N})(\frac{R_2}{N})(\frac{R_3}{N}) + (\frac{I_1}{N})(\frac{I_1}{N})(\frac{I_3}{N})
\label{eq:kappaeq}
\end{equation}
\begin{equation}
K = \frac{P(A) - P(E)}{1 - P(E)} 
\label{eq:kappaeq}
\end{equation} \\
\setcounter{table}{5}
\begin{table}[htb]
\centering
  \tbl{\caption{Results of Kappa inter-annotator agreement}}
 {
    \begin{tabular}{@{\extracolsep{\fill}}lcc}
    \hline
    Criteria & K & Agreement Level \\
    \hline
     Fluency & 74.1\% & Substantial\\ \hdashline
     Semantic & 75.32\% & Substantial\\ \hdashline
     Adulteration & 69.96\% & Substantial\\
    \hline
\end{tabular}}
    {\begin{tabnote}
    (SRC = Input Document; PGST is our method)
     \end{tabnote}}
  \label{tab:blplacc}
\end{table}
\-\hspace{0.40cm}K value is calculated for each facet in every group. Finally, their average score is stated in Table \hyperref[tab:kappatable]{7} which demonstrates the stability of annotations since such obtained results are counted as $substantial$ ones in Kappa inter-annotator agreement\textquotesingle{s} jargon.
\subsubsection{Quality Assessment}
Since our annotator groups each consist of three different annotators, when classifying each sample\textquotesingle{s} facet, we consider the two most agreed on opinion as the sample\textquotesingle{s} final class (e.g. if at least two out of three annotators classified a sample as 1 in fluency, we call that a fluent sample). Table \hyperref[tab:humantable]{8} demonstrates random samples\textquotesingle{} quality assessment based on their language and style.\\
\textbf{Fluency:} As mentioned in Algorithm \ref{alg:beam}, when choosing the right combination among all suggestions, we prioritize tokens with the highest frequency in different n-gram scopes (Equation \ref{eq:e7}) when choosing among replacements. This strategy will lead us towards a fluent $S_t$ document which is clearly proved by annotations shown in the table, with high accuracies of 77\% and 75\% in Persian and English. \\
\textbf{Semantic:} In terms of semantics, there are marginally lower accuracies obtained comparing to samples\textquotesingle{} overall fluency, probably mostly due to the literary essence of our Persian and informality essence of our employed English corpora, which is a two-faced scenario. On the one hand, it demonstrates how hazardous it is to find a decent replacement for intended tokens in such stylistic corpora. On the other hand, it indicates how compatible our method can be when applied to such different text. \\
\textbf{Adulteration:} The rationale for using $S_s$ documents in annotating process was to see if $S_t$ documents seemed evidently adulterated among the others, or that they obeyed of a similar format, which surprisingly, nearly the same amount of $S_t$ English documents were detected as adulterated as the $S_s$ English ones, which heralds of low difference among them. The relatively higher frequency of detected $S_t$ Persian documents, compared to $S_s$ Persian ones, is mostly because of poor replacement suggestions that our acquired pre-trained Persian word embedding gives (Since it is not as well trained as pre-trained English word vectors) and the cumbersome construction of Persian literary text, which is hard to cope with in transferring process. 
\setcounter{table}{6}
\begin{table}[htb]
\centering
  \tbl{\caption{Quality assessment of annotated samples}}
 {
    \begin{tabular}{@{\extracolsep{\fill}}lccc}
    \hline
    Set & Fluency & Semantic & Adulteration \\
    \hline
     $S_s$ Persian & 97.33\% & 97.33\% & 21.33\% \\ \hdashline
     $S_t$ Persian & 77.03\% & 63.51\% & 40.54\% \\ \hdashline
     $S_s$ English & 98.67\% & 90.67\% & 26.67\% \\ \hdashline
     $S_s$ English & 75.0\% & 68.42\% & 28.95\% \\
     \hline
    \end{tabular}}
  {\begin{tabnote}
  \end{tabnote}}
  \label{tab:humantable}
\end{table}
\setcounter{table}{7}
\begin{table}[htb]
\centering
  \tbl{\caption{Comparison of Automatic Evaluation results of different models in English}}
 {
    \begin{tabular}{@{\extracolsep{\fill}}lccc}
    \hline
    Model & BLEU & Perplexity & Accuracy \\
    \hline
     SRC & 100 & 183.4 & 18.9 \\ \hdashline
     BT & 46.0 & 196.2 & 52.9 \\ \hdashline
     G-GST & 78.5 & 252.0 & 49.0 \\ \hdashline
     B-GST & \textbf{82.5} & \textbf{189.2} & \textbf{57.9} \\
    \hline
     PGST & 68.4 & 198.9 & 45.6 \\
     \hline
    \end{tabular}}
    {\begin{tabnote}
    (SRC = Input Document; PGST is our method)
     \end{tabnote}}
  \label{tab:blplacc}
\end{table}
\subsection{Automatic Evaluation}
To indicate our proposed method\textquotesingle{s} correctness, we assess our method via automatic evaluation measurements in different aspects of fluency, content preservation and transfer strength. Previous works measured transfer strength with a style classifier, explicitly trained for evaluation purposes. But since proposing such a classifier was one of our key contributions, our model has already been particularized in previous sections. We employed BLEU \citep{bleupaper} and OpenAI GPT-2 language model respectively, to measuring content preservation and fluency of our transferred set. Although evaluating with such automatic metrics like BLEU is inadequate if being applied single-handedly \citep{badbleupaper}, it benefits us with a general understanding of how preserved a $S_t$ sample\textquotesingle{s} content is. But in fluency terms, we calculated the perplexity of those documents using GPT-2 language model.\\
\-\hspace{0.40cm}As a comparison with previous work, we intend to compare our method (\textbf{PGST}) on the analogy of three other previously proposed methods. \citep{btpaper} came up with the idea of adversarial mechanism and Back-Translation (\textbf{BT}) and \citep{sudhakar2019transforming} proposed \textbf{B-GST} and \textbf{G-GST} which respectively were blinded and guided towards particular desired $S_t$ style attributes using Transformers \citep{vaswani2017attention} which at the time of writing this paper and to our best of knowledge, is the state-of-the-art on our mutual English gender tagged dataset.
\-\hspace{0.40cm} When making an analogy, The key factor is not to consider each evaluation metric separately but to contemporaneously assess them all together. As shown in Table \hyperref[tab:blplacc]{9}, In terms of target style accuracy, the \textbf{BT} model performs admissibly well, but its generated text does not preserve much content, thus resulting in a low BLEU score, whereas in $S_t$ style matter our method almost obtained the same result as state-of-the-art method collateral, \textbf{G-GST}, but with much lower perplexity and higher BLEU score comparing to the prior state-of-the-art method, \textbf{BT}. All in all, it can be concluded that besides other monolingual methods, even with a much simpler foundation tp a multilingual extent, our proposed method has achieved reasonable success in English, whereas one of our main focuses was on devoting such a method in Persian.
\setcounter{table}{8}
\begin{table}[htb]
\centering
  \tbl{\caption{Test samples of transferring English text using the PGST method.}}
 {
\begin{tabular}{p{0.1\textwidth}p{0.4\textwidth}p{0.4\textwidth}}
\hline
Examples & $S_s$ style & $S_t$ style\\
\hline
EFM@1&they were very tasty, albeit surprisingly sweet!&they were extremely delicious, though quit sugary!\\
\hdashline
EFM@2&we arrived hours before they closed and we were said that they were not taking any more seating requests for the evening&we came hours before they shut and we were told that they were not taking any more seating requests for the night \\
\hdashline
EFM@3&it was a beautiful hotel and offers a variety of restaurants, shops, and services.&it was a great motel and provided a number of bistros, supermarkets, and service. \\
\hdashline
EFM@4&visually you will not step into many clubs in the world as beautiful as this place&perceptually you will never walk into countless clubs in the globe as lively as this site \\
\hdashline
EFM@5&my daughter got sick with fever after our recent visit&my wife became ill with pyrexia after our recent visit \\
\hline
\hline
Examples & $S_s$ style & $S_t$ style\\
\hline
EMF@1&brought some co-workers to lunch here and had a pretty good experience&brought some colleagues to lunch here and had a pretty amazing experience \\
\hdashline
EMF@2&you have to talk to the manager to understand whats going on to your bike&you have to discuss with the manager to figure whats going on with your bicycle \\
\hdashline
EMF@3&just horrible airline food and very limited selection of drinks!&just terrible airline meal and extremely limited selection of beverages! \\
\hdashline
EMF@4&most friendly and obliging staff and you never have to wait more than 30 mins since you get there&most friendly and accommodating personnel and you never have to wait more than 30 minutes when you get there \\
\hdashline
EMF@5&how good do you think it would taste to put cold veggies on warm bread&how nice do you believe it would taste to put cold broccoli on warm bread \\
\hline
\end{tabular}}
    {\begin{tabnote}
    Note: EFM stands for English Female to Male transfer. EMF stands for English Male to Female transfer.
    \end{tabnote}}
  \label{tab:blplacc}
\end{table}
\setcounter{table}{9}
\begin{table}[htb]
\centering
  \tbl{\caption{Test samples of transferring Persian text using the PGST method.}}
 {
\begin{tabular}{lrr}
\hline
Examples & $S_s$ style & $S_t$ style\\
\hline
PFM@1&
\includegraphics[scale=0.15]{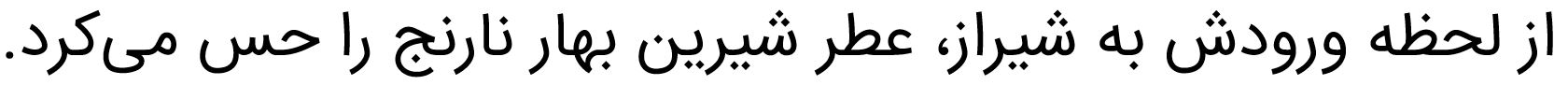}
&\includegraphics[scale=0.15]{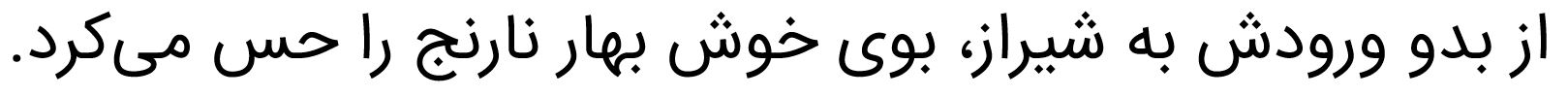} \\
\hdashline
PFM@2&
\includegraphics[scale=0.15]{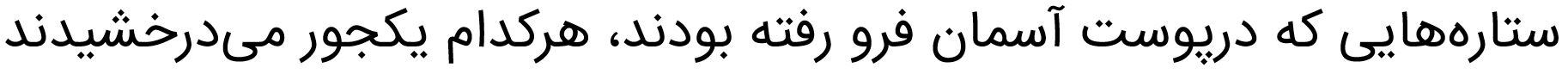}
&\includegraphics[scale=0.15]{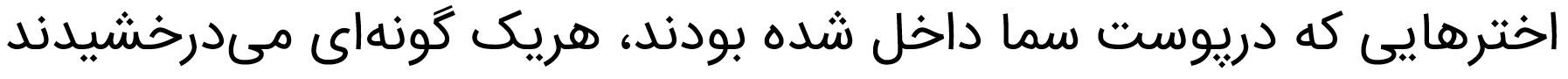} \\
\hdashline
PFM@3&
\includegraphics[scale=0.15]{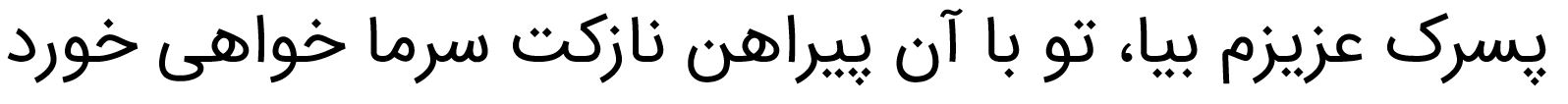}
&\includegraphics[scale=0.15]{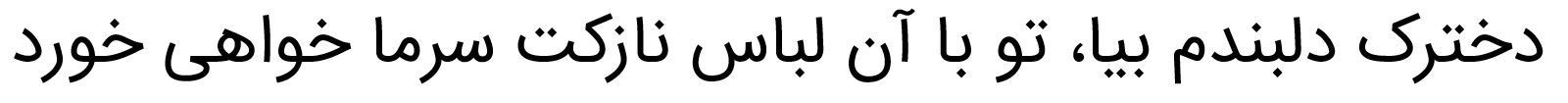} \\
\hdashline
PFM@4&
\includegraphics[scale=0.15]{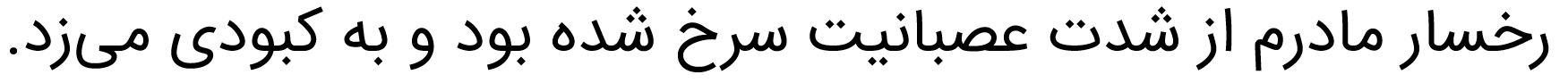}
&\includegraphics[scale=0.15]{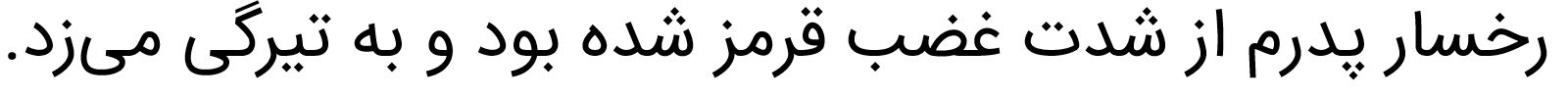} \\
\hdashline
PFM@5&
\includegraphics[scale=0.2]{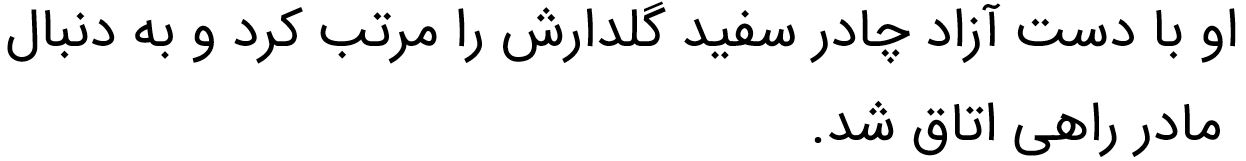}
&\includegraphics[scale=0.2]{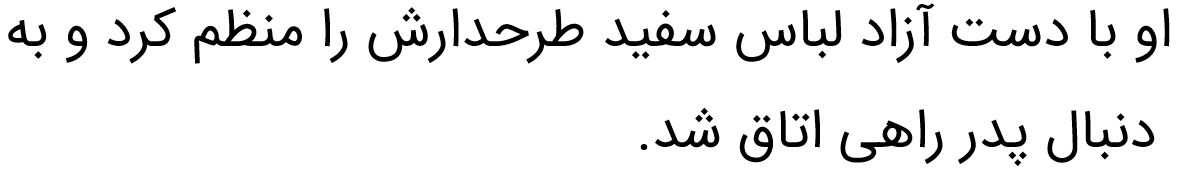} \\
\hline
\hline
Examples & $S_s$ style & $S_t$ style\\
\hline
PMF@1&
\includegraphics[scale=0.18]{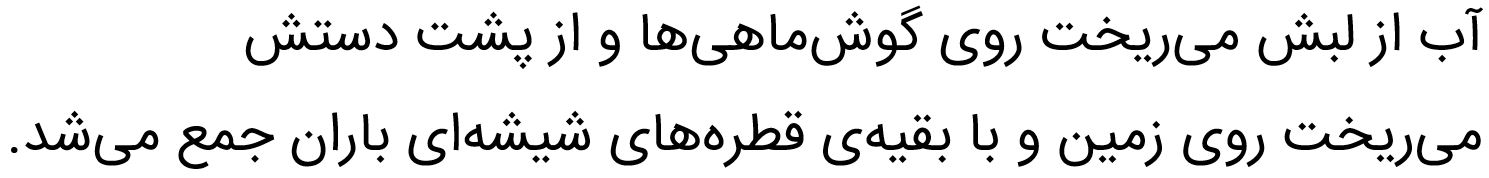}
&\includegraphics[scale=0.18]{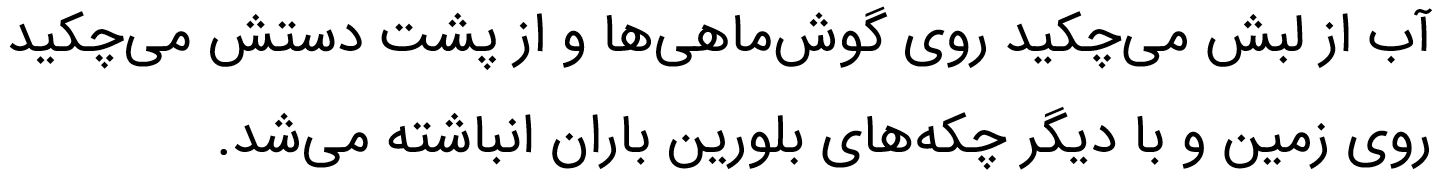} \\
\hdashline
PMF@2&
\includegraphics[scale=0.175]{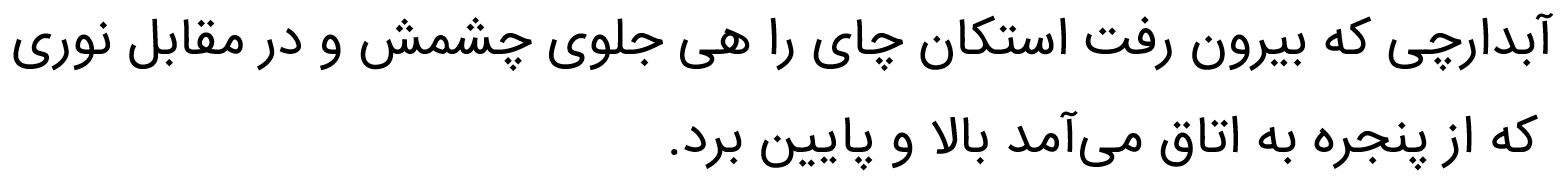}
&\includegraphics[scale=0.175]{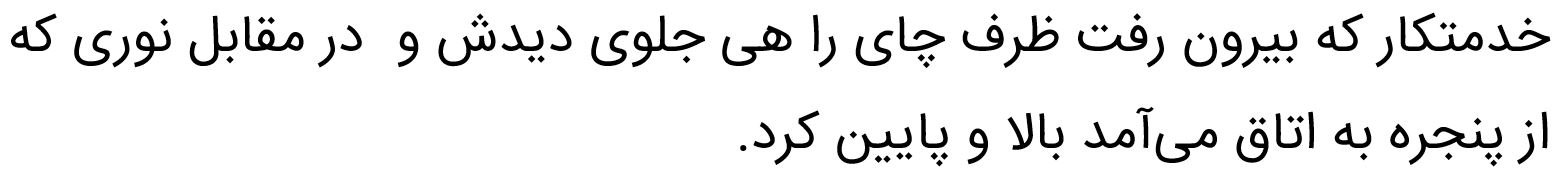} \\
\hdashline
PMF@3&
\includegraphics[scale=0.2]{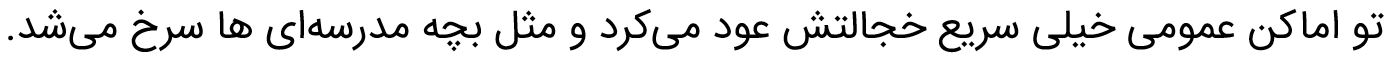}
&\includegraphics[scale=0.2]{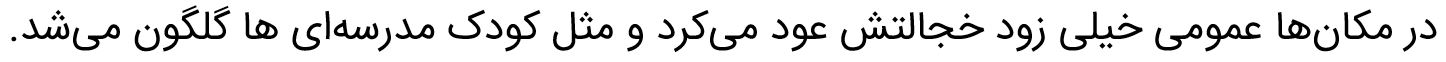} \\
\hdashline
PMF@4&
\includegraphics[scale=0.17]{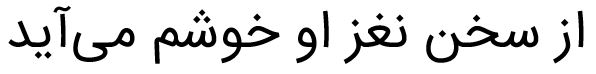}
&\includegraphics[scale=0.17]{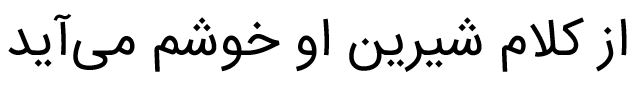} \\
\hdashline
PMF@5&
\includegraphics[scale=0.17]{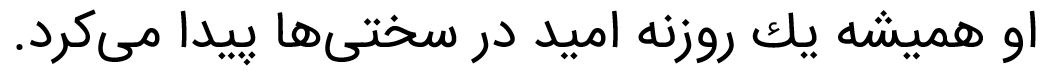}
&\includegraphics[scale=0.17]{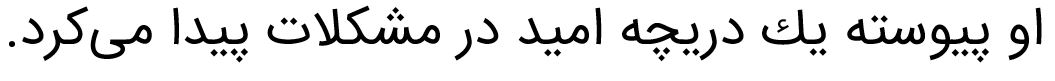} \\
\hline
\end{tabular}}
    {\begin{tabnote}
    Note: PFM stands for Persian Female to Male transfer. PMF stands for Persian Male to Female transfer.
    \end{tabnote}}
  \label{tab:blplacccc}
\end{table}
\setcounter{table}{10}
\begin{table}[htb]
\centering
  \tbl{\caption{Translation of Table \hyperref[tab:blplacccc]{11}'s test samples}}
 {
\begin{tabular}{p{0.1\textwidth}p{0.4\textwidth}p{0.4\textwidth}}
\hline
Translations & $S_s$ style & $S_t$ style\\
\hline
FM@1&From the moment he arrived in Shiraz, he could feel the sweet aroma of orange orange blossom&Upon his arrival in Shiraz, he smelled the pleasant aroma orange blossom.\\
\hdashline
FM@2&The stars that had plunged into the skin of the sky, each shone in a way&The stars that had entered the skin of the sky, each shone somehow\\
\hdashline
FM@3&Come on my dear son, you will catch a cold with that thin shirt of yours&Come on my beloved daughter, you will catch a cold with that thin dress of yours\\
\hdashline
FM@4&My mother's face was flushed red hot with anger and it got almost bruised-like.&My mother's face was flushed red with rage and it got almost dark.\\
\hdashline
FM@5&She organized her white veil With her free hand and followed her mother into the room.&He organized his white clothes With his free hand and followed his father into the room.\\
\hline
\hline
Translations & $S_s$ style & $S_t$ style\\
\hline
MF@1&Water poured from her lips on the scallops and poured from the back of her hand on the ground and got added to the rest of the glassy raindrops.&Water dripped from her lips on the scallops and dripped from the back of her hand on the ground and got accumulated with the rest of the crystalline raindrops.\\
\hdashline
MF@2&When the butler went out, she raised and lowered the cup of tea in front of her eyes and in front of the light that was coming from the window.&When the butler went out, she raised and lowered the cup of tea in front of her sight and in front of the light that was coming from the window.\\
\hdashline
MF@3&In public, she was quickly embarrassed and blushed like a schoolgirl.&In public, he was quickly embarrassed and reddened like a schoolboy.\\
\hdashline
MF@4&I like her bon mot&I like his sweet words\\
\hdashline
MF@5&She always found a glimmer of hope in hardships.&He constantly found a break of hope in difficulties.\\
\hline
\end{tabular}}
    {\begin{tabnote}
    Note: These Translations are not necessarily valid as English transferred samples and are only indicated to help non-Persian readers understand Table \hyperref[tab:blplacccc]{11} samples.
    \end{tabnote}}
  \label{tab:blplaccT}
\end{table}
\section{Discussion}
\label{discussion}
\-\hspace{0.40cm} In this section, we go over the advantages and disadvantages of our proposed method and study in what scenarios does the method succeed or fail to overcome the challenges it may face. We provide 10 test samples in both English (Table \hyperref[tab:blplacc]{10})) and Persian (Table \hyperref[tab:blplacccc]{11})). To depict the applicability of our method in transferring the textual style between the two genders, we dedicate 5 samples to both male to female and female to male text. \\
\-\hspace{0.40cm} It is noteworthy that besides their different languages, there are other significant differences between the two gender tagged datasets that we have employed in this research. The English dataset consists of reviews of different costumers on Yelp. On the other hand, the Persian dataset consists of different books and stories that have been written by either male or female authors. Hence, the model faces challenges that vary depending on the domain we apply it to. We describe the characteristics of each dataset below.
\newpage
\begin{itemize}
    \item The English dataset consists of mostly informal text, which makes it probable for the method to face OOV words unintentionally. Besides, producing fluent outputs becomes a considerable task since contractions (e.g. “can’t”), slang, abbreviations, and vague colloquial are often used in the informal text \citep{Fadi}.
    \item The Persian dataset, unlike English, consists of formal text. However, where the author aims to quote someone, we have spotted informal text. Altogether, the Persian dataset comprises mostly formal and few informal text each posing specific challenges. Such text follow a logical flow and complex structures. Like the English dataset, the vocabulary consists of OOV words, mostly due to the dataset's poetic or ancient literature. In contrast, almost no slangs, contractions, or abbreviations are seen. 
\end{itemize}
\-\hspace{0.40cm} Besides six middle eastern countries that partly speak it, the Persian/Farsi language is the official language of Iran, Afghanistan, and Tajikistan. Despite the Indo-European family of languages, Persian is one of the most important members of the Indo-Iranian branch. Persian has undergone significant changes as one of the most ancient languages and has been shaped over the years. Due to the adjacency of Persian and Arabic speakers, a plethora of Arabic loaned words have been injected into Persian. Since Persian speakers suggest replacements for such words and that there is no specified boundary on which Arabic words are officially accepted in Persian, the ratio of OOV words gets unintentionally even higher compared to English. As \citep{Mehrnoush} suggests, the following is a list of challenges that need to be handled to process this language: \\
\begin{enumerate}
\item	Generally, and most specifically at the lexical level, there is a vast gap between how the colloquial and formal Persian is spoken and written. Due to the change of grammar in colloquial, most resources  can only process the formal literature.
\item	To some extent, Persian has turned out to be free of word order. Meaning sentences are still expressive even if specific part-of-speech tags are relocated. However, such characteristic makes the language difficult to process, specifically in terms of Natural Language Understanding.
\item	There exist many scripts and types of writing a Persian letter.
\item	Unlike English, German, or French, Persian has no definite article.
\item	Uncountable nouns are probable to appear in plural form.
\item	Persian adjectives are likely to be used in place of nouns. This causes many semantic and structural ambiguities among noun phrases.
\item	Most related to our research, unlike English, Arabic, or German, there is no distinction between Male and Female nouns. 
\end{enumerate}
\-\hspace{0.40cm} Even though we have mentioned most of the critical entanglements that the Persian language’s essence poses, we refer readers to \citep{Mehrnoush} where they have dug deep in the topic. \\
\-\hspace{0.40cm} Having discussed the challenges of processing the Persian Language in an analogy to English, we consider interpreting our method's outcome. English examples (Table \hyperref[tab:blplacc]{10}) are either English Female to Male transfer (EFM) or English Male to Female transfer (EMF). The same is true for Persian Female to Male (PFM) and Persian Male to Female transfer (PMF). The $BeamSearch$ function's effect can be seen in EFM@5 and PFM@2, meaning that the expression has been previously seen in the literature. Besides, the PMF@1 sample indicates that Persian complex verbs are also being handled by the method. In specific terms, the method fails to ignore artificial replacements such as PFM@4, PFM@1 or "pyrexia" in EFM@5. Interestingly, even though Persian has no gender-distinguished nouns, "dress" and "clothes" are handled in Persian (PFM@3). Moreover, stemming and lemmatizing candidates may not necessarily aid us in selecting better replacements. Hence, we consider the word embeddings\textquotesingle{} suggestions in both filterd and raw formats of the word. For instance, PMF@1 and PMF@1 are examples that back this claim up.\\
\-\hspace{0.40cm} In brief, our method demonstrates that the PGST method succeeds in transferring the textual style not only in English but with an acceptable performance in Persian as well, which is not simple to be processed and to be coped with. Future work may focus on novel monolingual Text Style Transfer methods that take Persian-specific challenges into account.
\section{Conclusion and Future Work}
\label{conclusion}
This work introduced a novel polyglot approach for text style transfer, an essential modern task of natural language processing. Its primary focus is on transferring the style between the two male/female genders. Besides introducing the first instance of a Text Style Transfer method in Persian, we trained our model on English text to demonstrate its capability in other languages. We evaluated our method with statistical, automatic, and human metrics and explained how our obtained results became feasible from a sociolinguistic perspective. Results are highly competitive among robust English methods, and a baseline is set for future work in Persian Text Style Transfer.\\ 
\-\hspace{0.40cm} Unlike highly resourced natural languages in which text style transfer has turned out to be a well-developed task, more focused research is expected to be seen among low-resource languages. At the time of writing the paper, Attention-based models and Transformers are yet to be developed in Persian that can contribute to Persian significantly. \\

\bibliographystyle{nlelike}
\bibliography{mybibfile}

\begin{thebibliography}{}

\bibitem[Ahmadi et~al., 2016]{7585495}
{\bf Ahmadi, P.}, {\bf Tabandeh, M.}, \textbf{and} {\bf Gholampour, I.} 2016.
\newblock Persian text classification based on topic models.
\newblock In {\em 2016 24th Iranian Conference on Electrical Engineering
  (ICEE)}, pp. 86--91.

\bibitem[Brownlee, 2017]{brown}
{\bf Brownlee, J.} 2017.
\newblock {\em Deep Learning for Natural Language Processing, Machine Learning
  Mastery}.
\newblock Machine Learning Mastery.

\bibitem[Bsir and Zrigui, 2018]{bsir}
{\bf Bsir, B.} \textbf{and} {\bf Zrigui, M.} 2018.
\newblock Enhancing deep learning gender identification with gated recurrent
  units architecture in social text.
\newblock {\em Computacion y Sistemas}, 22:757–766.

\bibitem[Bucholtz, 2002]{Bucholtz}
{\bf Bucholtz, M.} 2002.
\newblock From 'sex differences' to gender variation in sociolinguistics.

\bibitem[Cavas, 2010]{Cavas2010ASO}
{\bf Cavas, B.} 2010.
\newblock A study on pre-service science, class and mathematics teachers'
  learning styles in turkey.
\newblock {\em Science education international}, 21:47--61.

\bibitem[Chen et~al., 2016]{chen2016infogan}
{\bf Chen, X.}, {\bf Duan, Y.}, {\bf Houthooft, R.}, {\bf Schulman, J.}, {\bf
  Sutskever, I.}, \textbf{and} {\bf Abbeel, P.} 2016.
\newblock Infogan: Interpretable representation learning by information
  maximizing generative adversarial nets.

\bibitem[Cheng et~al., 2011]{CHENG201178}
{\bf Cheng, N.}, {\bf Chandramouli, R.}, \textbf{and} {\bf Subbalakshmi, K.}
  2011.
\newblock Author gender identification from text.
\newblock {\em Digital Investigation}, 8(1):78--88.

\bibitem[Cheng et~al., 2020]{cheng2020contextual}
{\bf Cheng, Y.}, {\bf Gan, Z.}, {\bf Zhang, Y.}, {\bf Elachqar, O.}, {\bf Li,
  D.}, \textbf{and} {\bf Liu, J.} 2020.
\newblock Contextual text style transfer.

\bibitem[Devlin et~al., 2018]{devlin2018bert}
{\bf Devlin, J.}, {\bf Chang, M.-W.}, {\bf Lee, K.}, \textbf{and} {\bf
  Toutanova, K.} 2018.
\newblock Bert: Pre-training of deep bidirectional transformers for language
  understanding.

\bibitem[Eckert, 1989]{eck}
{\bf Eckert, P.} 1989.
\newblock The whole woman: Sex and gender differences in variation.
\newblock {\em Language Variation and Change}, 1:245--267.

\bibitem[Fatima et~al., 2018]{fatima2018}
{\bf Fatima, M.}, {\bf Anwar, S.}, {\bf Naveed, A.}, {\bf Arshad, W.}, {\bf
  Nawab, R. M.~A.}, {\bf Iqbal, M.}, \textbf{and} {\bf Masood, A.} 2018.
\newblock Multilingual sms-based author profiling: Data and methods.
\newblock {\em Natural Language Engineering}, 24(5):695–724.

\bibitem[Feng et~al., 2018]{inproceedings}
{\bf Feng, S.}, {\bf Wallace, E.}, {\bf II, A.}, {\bf Iyyer, M.}, {\bf
  Rodriguez, P.}, \textbf{and} {\bf Boyd-Graber, J.} 2018.
\newblock Pathologies of neural models make interpretations difficult.
\newblock pp. 3719--3728.

\bibitem[Fu et~al., 2017]{fu2017style}
{\bf Fu, Z.}, {\bf Tan, X.}, {\bf Peng, N.}, {\bf Zhao, D.}, \textbf{and} {\bf
  Yan, R.} 2017.
\newblock Style transfer in text: Exploration and evaluation.

\bibitem[Gatys et~al., 2016]{gatys2016preserving}
{\bf Gatys, L.~A.}, {\bf Bethge, M.}, {\bf Hertzmann, A.}, \textbf{and} {\bf
  Shechtman, E.} 2016.
\newblock Preserving color in neural artistic style transfer.

\bibitem[Gatys et~al., 2015]{gatys2015neural}
{\bf Gatys, L.~A.}, {\bf Ecker, A.~S.}, \textbf{and} {\bf Bethge, M.} 2015.
\newblock A neural algorithm of artistic style.

\bibitem[Gong et~al., 2019]{gong2019reinforcement}
{\bf Gong, H.}, {\bf Bhat, S.}, {\bf Wu, L.}, {\bf Xiong, J.}, \textbf{and}
  {\bf mei Hwu, W.} 2019.
\newblock Reinforcement learning based text style transfer without parallel
  training corpus.

\bibitem[Honnibal and Johnson, 2015]{honnibal-johnson-2015-improved}
{\bf Honnibal, M.} \textbf{and} {\bf Johnson, M.} 2015.
\newblock An improved non-monotonic transition system for dependency parsing.
\newblock In {\em Proceedings of the 2015 Conference on Empirical Methods in
  Natural Language Processing}, pp. 1373--1378, Lisbon, Portugal. Association
  for Computational Linguistics.

\bibitem[Hoyle et~al., 2019]{gdd}
{\bf Hoyle, A.}, {\bf Wolf-Sonkin}, {\bf Wallach, H.}, {\bf Augenstein, I.},
  \textbf{and} {\bf Cotterell, R.} 2019.
\newblock Unsupervised discovery of gendered language through latent-variable
  modeling.

\bibitem[Hu et~al., 2020]{hu2020text}
{\bf Hu, Z.}, {\bf Lee, R. K.-W.}, \textbf{and} {\bf Aggarwal, C.~C.} 2020.
\newblock Text style transfer: A review and experiment evaluation.

\bibitem[Hu et~al., 2017]{hu2017controlled}
{\bf Hu, Z.}, {\bf Yang, Z.}, {\bf Liang, X.}, {\bf Salakhutdinov, R.},
  \textbf{and} {\bf Xing, E.~P.} 2017.
\newblock Toward controlled generation of text.

\bibitem[Jin-yu, 2014]{Jinyu2014StudyOG}
{\bf Jin-yu, D.} 2014.
\newblock Study on gender differences in language under the sociolinguistics.
\newblock {\em Canadian Social Science}, 10:92--96.

\bibitem[John et~al., 2019]{john-etal-2019-disentangled}
{\bf John, V.}, {\bf Mou, L.}, {\bf Bahuleyan, H.}, \textbf{and} {\bf
  Vechtomova, O.} 2019.
\newblock Disentangled representation learning for non-parallel text style
  transfer.
\newblock In {\em Proceedings of the 57th Annual Meeting of the Association for
  Computational Linguistics}, pp. 424--434, Florence, Italy. Association for
  Computational Linguistics.

\bibitem[Joulin et~al., 2016]{arm2016bag}
{\bf Joulin, A.}, {\bf Grave, E.}, {\bf Bojanowski, P.}, \textbf{and} {\bf
  Mikolov, T.} 2016.
\newblock Bag of tricks for efficient text classification.

\bibitem[Jozefowicz et~al., 2016]{jozefowicz2016exploring}
{\bf Jozefowicz, R.}, {\bf Vinyals, O.}, {\bf Schuster, M.}, {\bf Shazeer, N.},
  \textbf{and} {\bf Wu, Y.} 2016.
\newblock Exploring the limits of language modeling.

\bibitem[Kayaoğlu, 2012]{articleMusta}
{\bf Kayaoğlu, M.} 2012.
\newblock Gender-based differences in language learning strategies of science
  students.
\newblock {\em Journal of Turkish Science Education}, 9.

\bibitem[Kim, 2015]{ttestpaper}
{\bf Kim, T.} 2015.
\newblock T test as a parametric statistic.
\newblock {\em Korean Journal of Anesthesiology}, 68:540.

\bibitem[Kingma and Ba, 2014]{kingma2014adam}
{\bf Kingma, D.~P.} \textbf{and} {\bf Ba, J.} 2014.
\newblock Adam: A method for stochastic optimization.

\bibitem[Kreyer, 2014]{vbn}
{\bf Kreyer, R.} 2014.
\newblock Baker, p. 2014. using corpora to analyze gender.
\newblock {\em International Journal of Corpus Linguistics}, 19.

\bibitem[Li et~al., 2019]{li2019domain}
{\bf Li, D.}, {\bf Zhang, Y.}, {\bf Gan, Z.}, {\bf Cheng, Y.}, {\bf Brockett,
  C.}, {\bf Sun, M.-T.}, \textbf{and} {\bf Dolan, B.} 2019.
\newblock Domain adaptive text style transfer.

\bibitem[Li et~al., 2018]{li-etal-2018-delete}
{\bf Li, J.}, {\bf Jia, R.}, {\bf He, H.}, \textbf{and} {\bf Liang, P.} 2018.
\newblock Delete, retrieve, generate: a simple approach to sentiment and style
  transfer.
\newblock In {\em Proceedings of the 2018 Conference of the North {A}merican
  Chapter of the Association for Computational Linguistics: Human Language
  Technologies, Volume 1 (Long Papers)}, pp. 1865--1874, New Orleans,
  Louisiana. Association for Computational Linguistics.

\bibitem[Li and Kirkup, 2007]{LI2007301}
{\bf Li, N.} \textbf{and} {\bf Kirkup, G.} 2007.
\newblock Gender and cultural differences in internet use: A study of china and
  the uk.
\newblock {\em Computers \& Education}, 48(2):301--317.

\bibitem[Luan et~al., 2017]{luan2017deep}
{\bf Luan, F.}, {\bf Paris, S.}, {\bf Shechtman, E.}, \textbf{and} {\bf Bala,
  K.} 2017.
\newblock Deep photo style transfer.

\bibitem[Ma, 2018]{mam}
{\bf Ma, E.} 2018.
\newblock Besides word embedding, why you need to know character embedding?

\bibitem[Martinc and Pollak, 2018]{Martinc2018ReusableWF}
{\bf Martinc, M.} \textbf{and} {\bf Pollak, S.} 2018.
\newblock Reusable workflows for gender prediction.
\newblock In {\em LREC}.

\bibitem[Martinc and Pollak, 2019]{martinc2019}
{\bf Martinc, M.} \textbf{and} {\bf Pollak, S.} 2019.
\newblock Combining n-grams and deep convolutional features for language
  variety classification.
\newblock {\em Natural Language Engineering}, 25(5):607–632.

\bibitem[Metin et~al., 2011]{METIN20112728}
{\bf Metin, M.}, {\bf Yılmaz, G.~K.}, {\bf Birişçi, S.}, \textbf{and} {\bf
  Coşkun, K.} 2011.
\newblock The investigating pre-service teachers’ learning styles with
  respect to the gender and grade level variables.
\newblock {\em Procedia - Social and Behavioral Sciences}, 15:2728--2732.
\newblock 3rd World Conference on Educational Sciences - 2011.

\bibitem[Mir et~al., 2019]{mir2019evaluating}
{\bf Mir, R.}, {\bf Felbo, B.}, {\bf Obradovich, N.}, \textbf{and} {\bf Rahwan,
  I.} 2019.
\newblock Evaluating style transfer for text.

\bibitem[Mohtaj et~al., 2018]{mohtaj-etal-2018-parsivar}
{\bf Mohtaj, S.}, {\bf Roshanfekr, B.}, {\bf Zafarian, A.}, \textbf{and} {\bf
  Asghari, H.} 2018.
\newblock {P}arsivar: A language processing toolkit for {P}ersian.
\newblock In {\em Proceedings of the Eleventh International Conference on
  Language Resources and Evaluation ({LREC} 2018)}, Miyazaki, Japan. European
  Language Resources Association (ELRA).

\bibitem[Moradi and Bahrani, 2016]{moradib}
{\bf Moradi, M.} \textbf{and} {\bf Bahrani, M.} 2016.
\newblock Automatic gender identification in persian text.

\bibitem[Nahavandi and Mukundan, 2014]{articlenahavandi}
{\bf Nahavandi, N.} \textbf{and} {\bf Mukundan, J.} 2014.
\newblock Language learning strategy use among iranian engineering efl
  learners.
\newblock {\em Advances in Language and Literary Studies}, 5:34--45.

\bibitem[Norberg, 2016]{nbv}
{\bf Norberg, C.} 2016.
\newblock Naughty boys and sexy girls: The representation of young individuals
  in a web-based corpus of english.
\newblock {\em Journal of English Linguistics}, 44:291--317.

\bibitem[Pant et~al., 2020]{inbook1112}
{\bf Pant, K.}, {\bf Verma, Y.}, \textbf{and} {\bf Mamidi, R.} 2020.
\newblock {\em SentiInc: Incorporating Sentiment Information into Sentiment
  Transfer Without Parallel Data}, pp. 312--319.

\bibitem[Papineni et~al., 2002]{bleupaper}
{\bf Papineni, K.}, {\bf Roukos, S.}, {\bf Ward, T.}, \textbf{and} {\bf Zhu,
  W.~J.} 2002.
\newblock Bleu: a method for automatic evaluation of machine translation.

\bibitem[Pearce, 2008]{Pearce1}
{\bf Pearce, M.} 2008.
\newblock Investigating the collocational behaviour of man and woman in the bnc
  using sketch engine.
\newblock {\em Corpora}, 3.

\bibitem[Prabhumoye et~al., 2018]{btpaper}
{\bf Prabhumoye, S.}, {\bf Tsvetkov, Y.}, {\bf Salakhutdinov, R.}, \textbf{and}
  {\bf Black, A.~W.} 2018.
\newblock Style transfer through back-translation.

\bibitem[Rao and Tetreault, 2018]{rao-tetreault-2018-dear}
{\bf Rao, S.} \textbf{and} {\bf Tetreault, J.} 2018.
\newblock Dear sir or madam, may {I} introduce the {GYAFC} dataset: Corpus,
  benchmarks and metrics for formality style transfer.
\newblock In {\em Proceedings of the 2018 Conference of the North {A}merican
  Chapter of the Association for Computational Linguistics: Human Language
  Technologies, Volume 1 (Long Papers)}, pp. 129--140, New Orleans, Louisiana.
  Association for Computational Linguistics.

\bibitem[Reddy and Knight, 2016]{reddy}
{\bf Reddy, S.} \textbf{and} {\bf Knight, K.} 2016.
\newblock Obfuscating gender in social media writing.
\newblock pp. 17--26.

\bibitem[Serizel and Giuliani, 2017]{olivaaaa}
{\bf Serizel, R.} \textbf{and} {\bf Giuliani, D.} 2017.
\newblock Deep-neural network approaches for speech recognition with
  heterogeneous groups of speakers including children.
\newblock {\em Natural Language Engineering}, 23(3):325–350.

\bibitem[Shamsfard, 2019]{Mehrnoush}
{\bf Shamsfard, M.} 2019.
\newblock Challenges and opportunities in processing low resource languages: A
  study on persian.

\bibitem[Sheikha and Inkpen, 2011]{Fadi}
{\bf Sheikha, F.} \textbf{and} {\bf Inkpen, D.} 2011.
\newblock Generation of formal and informal sentences.
\newblock pp. 187--193.

\bibitem[Shen et~al., 2017]{shen2017style}
{\bf Shen, T.}, {\bf Lei, T.}, {\bf Barzilay, R.}, \textbf{and} {\bf Jaakkola,
  T.} 2017.
\newblock Style transfer from non-parallel text by cross-alignment.

\bibitem[Soler~Company and Wanner, 2016]{company}
{\bf Soler~Company, J.} \textbf{and} {\bf Wanner, L.} 2016.
\newblock A semi-supervised approach for gender identification.

\bibitem[Sotelo et~al., 2020]{sotelo}
{\bf Sotelo, A.~F.}, {\bf G{\'o}mez-Adorno, H.}, {\bf Esquivel-Flores, O.},
  \textbf{and} {\bf Bel-Enguix, G.} 2020.
\newblock Gender identification in social media using transfer learning.
\newblock In {\bf Figueroa~Mora, K.~M.}, {\bf Anzurez~Mar{\'i}n, J.}, {\bf
  Cerda, J.}, {\bf Carrasco-Ochoa, J.~A.}, {\bf Mart{\'i}nez-Trinidad, J.~F.},
  \textbf{and} {\bf Olvera-L{\'o}pez, J.~A.}, editors, {\em Pattern
  Recognition}, pp. 293--303, Cham. Springer International Publishing.

\bibitem[Subramanian et~al., 2018]{s2018multipleattribute}
{\bf Subramanian, S.}, {\bf Lample, G.}, {\bf Smith, E.~M.}, {\bf Denoyer, L.},
  {\bf Ranzato, M.}, \textbf{and} {\bf Boureau, Y.-L.} 2018.
\newblock Multiple-attribute text style transfer.

\bibitem[Sudhakar et~al., 2019]{sudhakar2019transforming}
{\bf Sudhakar, A.}, {\bf Upadhyay, B.}, \textbf{and} {\bf Maheswaran, A.} 2019.
\newblock Transforming delete, retrieve, generate approach for controlled text
  style transfer.

\bibitem[Sulem et~al., 2018]{badbleupaper}
{\bf Sulem, E.}, {\bf Abend, O.}, \textbf{and} {\bf Rappoport, A.} 2018.
\newblock Bleu is not suitable for the evaluation of text simplification.

\bibitem[Sutskever et~al., 2014]{sutskever2014sequence}
{\bf Sutskever, I.}, {\bf Vinyals, O.}, \textbf{and} {\bf Le, Q.~V.} 2014.
\newblock Sequence to sequence learning with neural networks.

\bibitem[Trudgill, 1972]{tru}
{\bf Trudgill, P.} 1972.
\newblock Sex, covert prestige and linguistic change in the urban british
  english of norwich.
\newblock {\em Language in Society}, 1:179--195.

\bibitem[Vaswani et~al., 2017]{vaswani2017attention}
{\bf Vaswani, A.}, {\bf Shazeer, N.}, {\bf Parmar, N.}, {\bf Uszkoreit, J.},
  {\bf Jones, L.}, {\bf Gomez, A.~N.}, {\bf Kaiser, L.}, \textbf{and} {\bf
  Polosukhin, I.} 2017.
\newblock Attention is all you need.

\bibitem[Viera and Garrett, 2005]{kappapaper}
{\bf Viera, A.} \textbf{and} {\bf Garrett, J.} 2005.
\newblock Understanding interobserver agreement: The kappa statistic.
\newblock {\em Family medicine}, 37:360--3.

\bibitem[Wallentin, 2020]{WALLENTIN202081}
{\bf Wallentin, M.} 2020.
\newblock Chapter 6 - gender differences in language are small but matter for
  disorders.
\newblock In {\bf Lanzenberger, R.}, {\bf Kranz, G.~S.}, \textbf{and} {\bf
  Savic, I.}, editors, {\em Sex Differences in Neurology and Psychiatry},
  volume 175 of {\em Handbook of Clinical Neurology}, pp. 81 -- 102. Elsevier.

\bibitem[Yang et~al., 2018]{yang2018unsupervised}
{\bf Yang, Z.}, {\bf Hu, Z.}, {\bf Dyer, C.}, {\bf Xing, E.~P.}, \textbf{and}
  {\bf Berg-Kirkpatrick, T.} 2018.
\newblock Unsupervised text style transfer using language models as
  discriminators.

\bibitem[Yildiz, 2019]{istan}
{\bf Yildiz, T.} 2019.
\newblock A comparative study of author gender identification.
\newblock {\em TURKISH JOURNAL OF ELECTRICAL ENGINEERING \& COMPUTER SCIENCES},
  27:1052--1064.

\bibitem[Zare, 2013]{articlezare}
{\bf Zare, P.} 2013.
\newblock Exploring reading strategy use and reading comprehension success
  among efl learners.
\newblock {\em World Applied Sciences Journal}, 22:1566--1571.

\bibitem[Zhang et~al., 2015]{zhang2015characterlevel}
{\bf Zhang, X.}, {\bf Zhao, J.}, \textbf{and} {\bf LeCun, Y.} 2015.
\newblock Character-level convolutional networks for text classification.

\bibitem[Zhou et~al., 2020]{zhou2020exploring}
{\bf Zhou, C.}, {\bf Chen, L.}, {\bf Liu, J.}, {\bf Xiao, X.}, {\bf Su, J.},
  {\bf Guo, S.}, \textbf{and} {\bf Wu, H.} 2020.
\newblock Exploring contextual word-level style relevance for unsupervised
  style transfer.

\end{thebibliography}
\label{lastpage}
\end{document}